\documentclass{IEEEojcsys}
\usepackage[colorlinks,urlcolor=blue,linkcolor=blue,citecolor=blue]{hyperref}

\usepackage{array}
\usepackage[backend=biber, style=ieee, citestyle=numeric-comp]{biblatex}
\usepackage{textcomp}
\usepackage{lmodern}
\usepackage{parskip}

\usepackage{graphicx}
\usepackage{subfigure}
\usepackage{amsfonts}
\usepackage{amsmath}
\usepackage{booktabs}
\usepackage{multirow}
\usepackage{makecell}

\usepackage[nameinlink]{cleveref}
\crefname{problem}{Problem}{Problems}
\crefformat{problems}{Problems}
\crefname{constraint}{Constraint}{Constraints}
\crefformat{constraints}{Constraints}
\crefname{proposition}{Proposition}{Propositions}
\crefformat{propositions}{Propositions}

\usepackage{float}
\usepackage{pifont}
\newcommand{\cmark}{\ding{51}}%

\jvol{00}
\jnum{XX}
\paper{1234567}
\pubyear{2021}
\receiveddate{5 May 2025}
\reviseddate{25 July 2025}
\accepteddate{20 August 2025}
\publisheddate{9 September 2025}
\currentdate{22 October 2025}
\doiinfo{10.1109/OJCSYS.2025.3607845}

\newcommand{\anynorm}[1]{\left\lVert#1\right\rVert}
\newcommand{\norm}[1]{\left\lVert#1\right\rVert_2}

\usepackage[dvipsnames]{xcolor}

\usepackage{amsthm}
\newcommand*\diff{\mathop{}\!\mathrm{d}}
\newtheorem{theorem}{Theorem}

\let\labelindent\relax
\usepackage{enumitem} 
\newlist{properties}{enumerate}{1}
\setlist[properties,1]{
    label=\textbf{P\arabic*},
    ref=P\arabic*, 
    itemsep=0pt,  
    parsep=3pt    
}
\crefname{propertiesi}{property}{properties}
\Crefname{propertiesi}{Property}{Properties}

\newlist{hypotheses}{enumerate}{1}
\setlist[hypotheses,1]{
    label=\textbf{H\arabic*},
    ref=H\arabic*, 
    itemsep=0pt, 
    parsep=3pt  
}
\crefname{hypothesesi}{hypothesis}{hypotheses} 
\Crefname{hypothesesi}{Hypothesis}{Hypotheses}

\usepackage{nicefrac}
\usepackage{tikz}
\usetikzlibrary{arrows.meta,positioning, calc, shadows}
\usepackage[utf8]{inputenc}
\usepackage[T1]{fontenc}

\renewcommand\thesubsection{\thesection.\Alph{subsection}}
\crefname{subsection}{Section}{Sections}
\Crefname{subsection}{Section}{Sections}

\AtBeginEnvironment{appendices}{%
  \renewcommand\thesubsection{\Alph{subsection}}
  \crefname{subsection}{Appendix}{Appendices}
  \Crefname{subsection}{Appendix}{Appendices}
}

\begin{document}

\sptitle{Article Category}

\title{Leveraging Analytic Gradients in Provably Safe Reinforcement Learning}

\editor{This paper was recommended by Associate Editor F. A. Author.}

\author{Tim Walter\affilmark{1}, Hannah Markgraf\textsuperscript{\textdagger}\affilmark{1}, Jonathan K\"ulz\textsuperscript{\textdagger}\affilmark{1,2}, and Matthias Althoff\affilmark{1,2}}

\affil{Technical University of Munich, Department of Computer Engineering, 85748 Garching, Germany}
\affil{Munich Center for Machine Learning (MCML), 80538 Munich, Germany}

\corresp{CORRESPONDING AUTHOR: T. Walter (e-mail: \href{mailto:tim.walter@tum.de}{tim.walter@tum.de})}
\authornote{This work was supported by the Deutsche Forschungsgemeinschaft (German Research Foundation) under grant numbers AL 1185/31-1 and AL 1185/9-1. The authors gratefully acknowledge the computational and data resources provided by the Leibniz Supercomputing Centre (www.lrz.de).}

\markboth{PREPARATION OF PAPERS FOR IEEE OPEN JOURNAL OF CONTROL SYSTEMS}{F. A. Author {\itshape ET AL}.}

\begin{abstract}
  The deployment of autonomous robots in safety-critical applications requires safety guarantees. Provably safe reinforcement learning is an active field of research that aims to provide such guarantees using safeguards. These safeguards should be integrated during training to reduce the sim-to-real gap. While there are several approaches for safeguarding sampling-based reinforcement learning, analytic gradient-based reinforcement learning often achieves superior performance from fewer environment interactions. However, there is no safeguarding approach for this learning paradigm yet. Our work addresses this gap by developing the first effective safeguard for analytic gradient-based reinforcement learning. We analyse existing, differentiable safeguards, adapt them through modified mappings and gradient formulations, and integrate them into a state-of-the-art learning algorithm and a differentiable simulation. Using numerical experiments on three control tasks, we evaluate how different safeguards affect learning. The results demonstrate safeguarded training without compromising performance. Additional visuals are provided at \href{https://timwalter.github.io/safe-agb-rl.github.io}{timwalter.github.io/safe-agb-rl.github.io}.
\end{abstract}

\begin{IEEEkeywords}
  Safe reinforcement learning, policy optimisation, differentiable simulation, gradient-based methods, constrained optimisation, first-order analytic gradient-based reinforcement learning
\end{IEEEkeywords}

\maketitle
\def\thefootnote{\textdagger}\footnotetext{Equal contribution}\def\thefootnote{\arabic{footnote}}

\section{INTRODUCTION}
The transfer of physical labour from humans and human-operated machines to robots is a long-standing goal of robotics research. Although robots have been successfully deployed in controlled environments, such as factories, their deployment in human proximity remains challenging \cite{StrongRobot2}. One reason is a lack of safety guarantees to ensure that robots do not harm humans or themselves~\cite{SafetyRequired}.
\par
A fundamental requirement for safe human-robot interaction is the deployment of controllers with provable safety guarantees. This becomes particularly challenging when using reinforcement learning, which often outperforms classical control methods in uncertain, high-dimensional, and non-linear environments \cite{GreatRL, GreatRL2}. To avoid costly and slow real-world training, an agent should preferably train in simulation before deployment to real systems \cite{Real1, Real2}. If the system is safety-critical, applying safeguards already during training is desirable to reduce the sim-to-real gap \cite{simtoreal1, simtoreal2}. Otherwise, the unsafeguarded optimisation may converge to a policy that relies on unsafe states or actions. When the deployed safeguard subsequently restricts the policy, it can become suboptimal or even fail for non-convex objective landscapes, as it has not learned alternative, safe solutions \cite{DuringTraining}. While crafting reward functions that reliably encode safety requirements could theoretically prevent performance degradation during deployment, this is notoriously difficult without introducing unintended incentives \cite{RewardHacking1, RewardHacking2}. Moreover, safeguards can aid learning by guiding exploration in challenging solution spaces \cite{RayMasking, PostTraining}.
\par
In recent years, provably safe reinforcement learning has emerged as a research field \cite{ProvablySafeSurvey, ProvablySafeRL}. Current safeguards are applied in conjunction with reinforcement learning algorithms that rely on the policy gradient theorem to estimate reward landscapes \cite{ActionReplacement1, ActionProjection, RayMasking, BoundaryProjectionLoss, Gauge, BoundaryProjection}. The advent of differentiable physics simulators \cite{Mujoco, brax, ChainQueen, Thuerey, SAPO} eliminates the need for this estimation, as a differentiable simulator enables the analytical computation of the reward gradient with respect to actions by backpropagation through the dynamics. While these simulators require approximations to remain differentiable, maintaining simulation accuracy is possible. Reinforcement learning algorithms that exploit these gradients promise faster training and better performance \cite{GradientRL1, GradientRL2, PODS, SHAC}. However, existing safeguards for sampling-based reinforcement learning can not be naively applied to these algorithms. Furthermore, there are currently no safeguarding mechanisms tailored to analytic gradient-based reinforcement learning.
\par
Our work combines state-of-the-art analytic gradient-based reinforcement learning algorithms, differentiable safeguards, and differentiable simulations. As differentiable safeguards, we incorporate a range of provably safe set-based safeguarding methods, whose codomain is a subset of a verified safe action set. By construction, this set consists only of safe actions. We formalise desirable safeguarding properties in the context of differentiable optimisation and analyse existing methods with respect to these criteria. Based on this analysis, we propose targeted modifications, such as custom backward passes or adapted maps, that enhance the suitability for analytic gradient-based reinforcement learning. We also extend the applicability of one of the safeguards to state constraints.
\par
We evaluate the provably safe approaches in differentiable simulations of various control problems. We observe sample efficiency and final performance that exceeds or is on par with unsafe training and sampling-based baselines.
\par
In summary, our core contributions are:
\begin{itemize}
  \item the first provably safe policy optimisation approach from analytic gradients\footnote{Code available at \href{https://github.com/TimWalter/SafeGBPO}{github.com/TimWalter/SafeGBPO}};
  \item an in-depth analysis of some suitable safeguards;
  \item adapted backward passes, an adapted mapping, and extended applicability of these safeguards to state constraints; and
  \item an evaluation on three control tasks, demonstrating the potential of provably safe reinforcement learning from analytic gradients.
\end{itemize}

\section{RELATED WORK}
We provide a literature review on the most relevant research areas: analytic gradient-based reinforcement learning, safeguards, and implicit layers that realise computing analytical gradients for optimisation-based safeguards.

\subsection{ANALYTIC GRADIENT-BASED REINFORCEMENT LEARNING}
Analytic gradient-based reinforcement learning relies on a continuous computational graph from policy actions to rewards, which allows computing the first-order gradient of the reward with respect to the action via backpropagation. Relying on first-order gradient estimators often results in less variance than zeroth-order estimators \cite{GradientOrderDoubt}, which are usually obtained using the policy gradient theorem. Less variance leads to faster convergence to local minima of general non-convex smooth objective functions \cite{GradientRL1, GradientRL2}. However, complex or contact-rich environments may lead to optimisation landscapes that are stiff, chaotic, or contain discontinuities, which can stifle performance as first-order gradients suffer from empirical bias \cite{GradientOrderDoubt}. Using a smooth surrogate to approximate the underlying noisy reward landscape can alleviate this issue \cite{SHAC}. Moreover, naively backpropagating through time \cite{BPTT} can lead to vanishing or exploding gradients in long trajectories \cite{BPTTProblems}.
\par
Various approaches have been introduced to overcome this issue: Policy optimisation via differentiable simulation \cite{PODS} utilises the gradient provided by differentiable simulators in combination with a Hessian approximation to perform policy iteration, which outperforms sampling-based methods. Short-horizon actor-critic (SHAC) \cite{SHAC} tackles the empirical bias of first-order gradient estimators by training a smooth value function through a mean-squared-error loss, with error terms calculated from the sampled short-horizon trajectories through a TD-$\lambda$ formulation \cite{RLBible}. It prevents exploding and vanishing gradients by cutting the computational graph deliberately after a fixed number of steps and estimating the terminal value by the critic. The algorithm shows applicability even in contact-rich environments, which tend to lead to stiff dynamics. The successor adaptive horizon actor-critic \cite{AHAC} has a flexible learning window to avoid stiff dynamics and shows improved performance across the same tasks. Short-horizon actor-critic also inspired soft analytic policy optimisation \cite{SAPO}, which integrates maximum entropy principles to escape local minima.

\subsection{SAFEGUARDING REINFORCEMENT LEARNING}
Safeguards are generally categorised according to their safety level \cite{Categorisation, ProvablySafeRL}. Since we seek guarantees, we limit the discussion to hard constraints. Moreover, safeguards for analytic gradient-based reinforcement learning must define a differentiable map from unsafe to safe actions to allow for backpropagation.
\par
Within this field of research, two common approaches for enforcing safety guarantees are control barrier functions \cite{cheng2019end, yang2020safe, marvi2022reinforcement, xiao2023safe, kokolakis2022safety} and reachability analysis \cite{selim2022safe, eichelbeck2022contingency, ActionProjection}. Both necessitate some form of environment model in their basic form, which can be identified from data \cite{wabersich2023data, Reachability, lutzow2024reachset}. By finding a control barrier function for a given system, forward invariance of a safe state set can be guaranteed \cite{cheng2019end}. While mainly used for control-affine systems, solutions for non-affine systems that rely on trainable high-order control barrier functions exist \cite{xiao2023safe}. Nevertheless, finding suitable candidates for control barrier functions for complex systems is non-trivial, and uncertainty handling remains challenging \cite{wabersich2023data}. Therefore, we employ reachability analysis, which uses non-deterministic models that capture the actual environment dynamics, to compute all possible system states \cite{ActionProjection}. Containment of the reachable state set in a safe state set can be guaranteed by adjusting reinforcement learning actions via constrained optimisation. If robust control invariant sets \cite{RCI-Sets} and reachset-conformant system identification are used \cite{lutzow2024reachset}, this approach can be applied efficiently to non-affine systems with uncertainties.
\par
Differentiable maps between unsafe and safe actions are required to combine the safeguarding approaches with analytic gradient-based reinforcement learning, which are only available for continuous action spaces. Krasowski et al. \cite{ProvablySafeRL} present continuous action projection with safe action sets represented by intervals, where straightforward re-normalisation is employed to map from the feasible action set. Stolz et al. \cite{RayMasking} generalise this to more expressive sets with their ray mask method. Tabas et al. \cite{Gauge} derive a differentiable bijection based on Minkowski functionals and apply it to power systems. Chen et al. \cite{BoundaryProjectionLoss} define differentiable projection layers relying on convex constraints. Gros et al. \cite{BoundaryProjection} define the mapping as an optimisation problem to determine the closest safe action. While these approaches are, in principle, differentiable, previous work only utilises them to modify policy gradients. In particular, we utilise and modify boundary projection \cite{BoundaryProjection} and ray masking \cite{RayMasking} to modify policy behaviour in a differentiable setting.

\subsection{IMPLICIT LAYERS} Defining the safeguards above can often not be done in closed form. Instead, they can only be formulated implicitly as a separate optimisation problem. Implicit layers \cite{ImplicitLayers, CVXPY-Layers} enable an efficient backpropagation through the solution of this separate optimisation problem without unrolling the solver steps. They decouple the forward and backward pass and analytically differentiate via the implicit function theorem \cite{ImplicitFunctionTheorem} using only constant training memory. Implicit layers are a potent paradigm that can be utilised for the tuning of controller parameters \cite{CVXPY-Control}, model identification \cite{CVXPY-Models}, and safeguarding \cite{BoundaryProjectionLoss}. Given the complexity of general optimisation problems being NP-hard, it is crucial to approach the implicit formulation with diligence. If a restriction to convex cone programs is possible, solutions can be computed efficiently in polynomial time \cite{diffcp}, thereby facilitating a swift forward pass \cite{GeometricProgramming}\cite{ConvexConicPrograms}. Moreover, this enables formulating the problem with CVXPY \cite{CVXPY, CVXPY2}, which automatically picks an efficient solver and translates the problem to the desired solver formulation.

\section{PRELIMINARIES}
We briefly introduce reinforcement learning based on analytical gradients, safe action sets, which serve as the notion of provable safety throughout this work, and zonotopes as a set representation for safe action sets.

\subsection{ANALYTIC GRADIENT-BASED REINFORCEMENT LEARNING}
Traditionally, deep reinforcement learning learns an action policy based on scalar rewards without assuming access to a model of the environment dynamics. Prominent algorithms such as REINFORCE~\cite{Williams1992}, proximal policy optimisation~\cite{PPO}, or soft actor-critic~\cite{SAC}, are based on the policy gradient theorem. This theorem provides a zeroth-order estimator for the gradient of the expected return $J(\theta)=\mathbb{E}_{\pi_\theta}\left\{\sum_{t=0}^T r(s_t, a_t)\right\}$, where $\pi_\theta$ is the parameterised policy, with respect to the policy parameters $\theta$, given by \cite[Eq. 2]{Sutton1999}:
\begin{align*}
  \frac{\partial J(\theta)}{\partial \theta}  = \mathbb{E}_{\pi_\theta} \left[\frac{\partial}{\partial \theta} \log \pi_\theta (a \mid s ) Q^\pi(s, a) \right] \, ,
\end{align*}
where $Q^\pi(s_t, a_t)$ denotes the action-value function under policy $\pi_\theta$. This gradient estimate can be used to optimise the policy via stochastic gradient descent.
\par
In contrast, analytical gradient-based reinforcement learning aims to replace this sample-based estimator with a direct gradient computed through a differentiable model of the environment. In such cases, the chain rule can be applied to the entire reward computation, yielding an analytical first-order estimate of the policy gradient:
\begin{align*}
  \frac{\partial J(\theta)}{\partial \theta} = \sum_{t=0}^T \left( \frac{\partial r(s_t, a_t)}{\partial s_t}\frac{\partial s_t}{\partial \theta} + \frac{\partial r(s_t, a_t)}{\partial a_t} \frac{\partial a_t}{\partial \theta} \right) \, ,
\end{align*}
where we use the numerator layout, i.e., the row number of $\frac{\partial y}{\partial x}$ equals the size of the numerator $y$ and the column number equals the size of $x^T$, for gradients throughout the paper. The term $\frac{\partial s_t}{\partial \theta}$ requires backpropagation through time, which can become numerically unstable for long trajectories. This problem motivates the introduction of a regularising critic, resulting in the short-horizon actor-critic algorithm~\cite{SHAC}. For a more detailed review of reinforcement learning methods based on analytical gradients, we refer the interested reader to~\cite{SHAC, AHAC, SAPO}.

\subsection{SAFE ACTION SETS} \label{ssec:safety}
To achieve provable safety, the safety of all traversed states and executed actions must be verifiable. Thus, we introduce a subset of the feasible state set $\mathcal{S}$, the safe state set $\mathcal{S}_s \subseteq \mathcal{S}$, containing all states that fulfil all safety specifications. Furthermore, we assume that provable safety is in principle possible, i.e., starting from a safe state, there must exist a sequence of safe actions that ensures the safety of all traversed states \\\cite[Proposition 1]{ProvablySafeRL}
\begin{equation}
  \forall s_0 \in \mathcal{S}_s \, \exists (a_0, a_1,\dots) \, \forall i \in \mathbb{N}: \mathcal{S}_{i+1}(a_i, s_i) \subseteq \mathcal{S}_s \, ,
  \label{proposition:existence}
\end{equation}
where $\mathcal{S}_{i+1}(a_i, s_i) $ denotes the next state set, i.e., the set of reachable states when executing action $a_i$ in safe state $s_i$. Given \Cref{proposition:existence}, there exists a non-empty safe action set
\begin{equation}
  \mathcal{A}_s(s_i) = \left\{a_i \in \mathcal{A} \mid \mathcal{S}_{i+1}(a_i, s_i) \subseteq \mathcal{S}_{s} \right\}
  \label{eq:safe_action_set}
\end{equation}
from which a policy can select actions. We refer to the safe action set in \Cref{eq:safe_action_set} as a \textit{derived} safe action set, since it is derived from the underlying state constraints $\mathcal{S}_s$, in contrast to a \textit{specified} safe action set, which may be defined directly. In this work, we introduce safeguards $g: \mathcal{X} \mapsto \mathcal{Y}$ with domain $\mathcal{X} \supseteq \mathcal{A}$ and codomain $\mathcal{Y} \subseteq \mathcal{A}_s$ that map any feasible, policy-selected action $a_i \in \mathcal{A}$ to a safe action $g(a_i) = a_{s,i} \in \mathcal{A}_s$. These safeguards are therefore provably safe by construction.

\subsection{ZONOTOPES}
We use zonotopes to represent safe sets due to their compact representation and closedness under linear maps and Minkowski sums. Zonotopes are convex, restricted polytopes and are defined as \cite[Eq. 3]{kuhnRigorouslyComputedOrbits1998}
\begin{equation}
  \mathcal{Z} = \left\{c+G\beta \mid \Vert{\beta}\Vert_\infty \leq 1 \right\} = \left<c, G\right>\,
\end{equation}
with centre $c \in \mathbb{R}^d$, generator matrix $G \in \mathbb{R}^{d \times n}$, and scaling factors $\beta \in [-1, 1]^n$. Zonotopes with orthogonal generators and $d=n$ are boxes. We utilise the following properties of zonotopes to formulate our safeguards. The Minkowski sum of two zonotopes $\mathcal{Z}_1,\mathcal{Z}_2 \subset \mathbb{R}^d$ is \cite[Eq. 7a]{RCI-Sets}
\begin{equation}
  \mathcal{Z}_1 \oplus \mathcal{Z}_2 = \left<c_1 + c_2, \begin{bmatrix} G_1 & G_2 \end{bmatrix}\right> \, . \label{eq:minkowski_sum}
\end{equation}
Translating a zonotope is equivalent to translating the centre. Linearly mapping by $M \in \mathbb{R}^{m \times d}$ yields \cite[Eq. 7b]{RCI-Sets}
\begin{equation}
  M\mathcal{Z} = \left<Mc, MG\right> \, .
  \label{eq:mapping}
\end{equation}
A support function of a set describes the farthest extent of the set in a given direction. The support function of a zonotope in direction $v \in \mathbb{R}^d$ is \cite[Lemma 1]{SupportFunction}
\begin{equation}
  \rho_{\mathcal{Z}}(v) = v^Tc+\anynorm{G^Tv}_1 \, .
\end{equation}
A point $p \in \mathbb{R}^d$ is contained in a zonotope if \cite[Eq. 6]{zonotopePointContainment}
\begin{equation}
  1 \geq \min_{\gamma\in\mathbb{R}^n} \Vert\gamma\Vert_\infty\, \text{s.t.}\, p = c + G\gamma \, . \label{eq:point_containment}
\end{equation}
Determining the containment of a zonotope in another zonotope is co-NP complete \cite{zonotopePointContainment}, but a sufficient condition for $\mathcal{Z}_1 \subseteq \mathcal{Z}_2$ is \cite[Eq. 15]{zonotopeContainment}
\begin{subequations}
  \label{eq:set_containment}
  \begin{alignat}{2}
    1 \geq & \!\min_{\gamma \in \mathbb{R}^{n_2}, \Gamma \in \mathbb{R}^{n_2 \times n_1}} & \qquad & \anynorm{\begin{bmatrix} \Gamma & \gamma \end{bmatrix}}_\infty \\
           & \text{subject to}                                                            &        & G_1 = G_2 \Gamma                                               \\
           &                                                                              &        & c_2 - c_1 = G_2 \gamma \, .
  \end{alignat}
\end{subequations}
Both containment problems are linear.

\section{PROBLEM STATEMENT} \label{sec:problem_statement}
Our work considers constrained Markov decision processes $(\mathcal{S}, \mathcal{A}, P_f, r, \mathcal{A}_s)$ with the following elements:

\begin{itemize}
  \item a feasible state set $\mathcal{S} \subseteq \mathbb{R}^{d_\mathcal{S}}$,
  \item a feasible action set $\mathcal{A} \subseteq \mathbb{R}^{d}$,
  \item a transition distribution $P_f(s_{i+1} |s_i, a_i)$,
  \item a continuously differentiable reward function $r(s_i, a_i, s_{i+1}) = r_i$,
  \item and a safe action set $\mathcal{A}_s \subseteq \mathcal{A}$.
\end{itemize}
We seek a safeguarded, stochastic policy that maximises the expected, discounted return over a finite horizon $N$:
\begin{equation}
  \pi^*(a|s) = \!\operatorname*{argmax}_{\pi(a|s)} \operatorname*{\mathbb{E}}_{\substack{a_i \sim \pi(a_i|s_i) \\ s_{i+1} \sim P_f}} \sum_{i=0}^N \delta^i \, r(s_i, a_{s,i}, s_{i+1})
\end{equation}
with the safe action $a_{s,i} = g(a_i)$, the continuously differentiable safeguard $g \colon \mathcal{A} \to \mathcal{A}_s$, and discount factor $\delta \in (0,1]$.

\subsection{ENSURING COMPUTATIONAL TRACTABILITY}
We impose additional requirements on the problem to ease the computational burden of safeguarding. This concerns the representation of all sets as closed, convex sets, such as zonotopes.
\par
Safeguarding is computationally cheap when the problem setting provides a specified safe action set. However, safeguarding might be computationally intractable if the safe action set needs to be constructed from a safe state set, as specified in \Cref{eq:safe_action_set}, even if the safe state set is available as a closed, convex set. In these cases, we assume that the next state set can be derived using disciplined convex programming \cite{dcp}. This makes it possible to replace any constraint on a safe action $a_{s,i} \in \mathcal{A}_s$ by the state constraint $\mathcal{S}_{i+1}(a_{s,i}, s_i) \subseteq \mathcal{S}_s$.
\par
In practice, the next state set is often enclosed via reachability analysis, leading to a conservative under-approximation of the safe action set in the current state. However, maintaining \Cref{proposition:existence} requires tight enclosures, which is an active field of research for complex systems \cite{Reachability, Overapproximation, Overapproximation1, RCI-Sets, setRA, aroc}.
\par
One method we discuss in particular is obtaining safe state sets via robust control invariant sets \cite{RCI-Sets}, which guarantee the existence of an invariance-enforcing controller that can keep all future states within the safe set. This is achieved by enclosing the dynamics \textit{at the current state} by a linear transition function with a noise zonotope $\mathcal{W} = \left<c_\mathcal{W}, G_\mathcal{W}\right> \subset \mathbb{R}^{d_\mathcal{S}}$, such that:
\begin{equation}
  \mathcal{S}_{i+1}(a_i, s_i) = M a_i \oplus \left<c + c_\mathcal{W}, G_{\mathcal{W}}\right> \, ,
  \label{eq:next_state_set}
\end{equation}
where $c$ is the offset and $M$ the Jacobian of the linearisation. Such an enclosure can, for example, be obtained using reachset-conformant identification \cite{Reachability}.

\section{METHOD} \label{sec:method}
\begin{figure}
  \centering
  \includegraphics[width=0.5\textwidth]{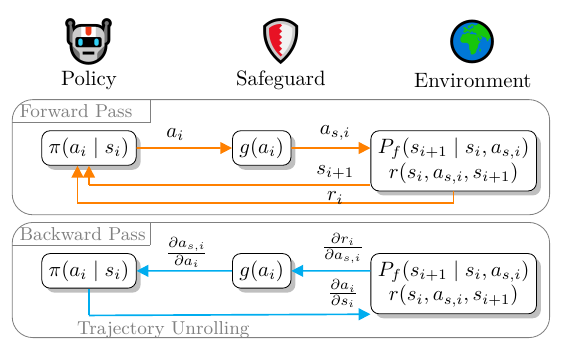}
  \caption{
    The forward (top) pass of the provably safe policy optimisation from analytic gradients describes the integration of the safeguard between the policy and environment. The backward pass (bottom) visualises how we utilise backpropagation to obtain the reward gradient with respect to the policy action. It also highlights the required unrolling of the previous trajectory.}
  \label{fig:pipeline}
\end{figure}

\Cref{fig:pipeline} shows the general framework for provably safe, analytic gradient-based reinforcement learning. For any policy output, we apply safeguards that map the unsafe action $a_i$ to the safe action $a_{s,i}$. The safe action is executed in the environment, yielding the next state $s_{i+1}$ and reward $r_{i}$. To train the policy, we calculate the gradient of the reward with respect to the policy output as:
\begin{equation}
  \frac{\partial r_i}{\partial a_i} = \left(\underbrace{\frac{\partial r_i}{\partial a_{s,i}}}_{\text{direct path}} + \underbrace{\frac{\partial r_i}{\partial s_{i+1}} \frac{\partial s_{i+1}}{\partial a_{s,i}} }_{\text{indirect path via }s_{i+1}}\right)\frac{\partial a_{s,i}}{\partial a_i} \, .
\end{equation}
Since the policy output and reward depend on the previous state, full backpropagation requires unrolling the trajectory to determine how all previous policy outputs affect the current reward.
\par
The following subsections detail the safeguard. First, we formulate generally required and desirable properties in the aforementioned differentiable setting. Then, we introduce the two safeguards used in this work: boundary projection (BP) \cite{BoundaryProjection} and ray mask (RM) \cite{RayMasking}. We structure their introduction by first explaining the general idea of the safeguard, then analysing its properties, and finally presenting our modifications.

\subsection{REQUIRED AND DESIRED PROPERTIES}
A safeguard for our setting must be provably safe as described in \Cref{ssec:safety}. For backpropagation, it must also guarantee the existence of a Clark generalised derivative \cite{Clarke} everywhere, which implies that the safeguarding is at least of class $C^0$.
\par
Beyond the required properties, there are additional desired properties. First, the safeguarding should be of class $C^1$ and provide full rank Jacobians $\frac{\partial a_s}{\partial a}$ everywhere. A rank-deficient Jacobian can diminish the learning signal by reducing its effective dimensionality, which incurs information loss. Second, the number of interventions by the safeguarding should be minimal throughout training and inference. Minimal interventions also reduce overhead and serve the last desired property of fast computation. In summary, the safeguard \textit{must}
\begin{properties}
  \item map any action to a safe action, \label{prop:1}
  \item be subdifferentiable everywhere, and therefore of class $C^0$  \label{prop:2}
\end{properties}
and \textit{should}
\begin{properties}[start=3]
  \item be of class $C^1$ and provide full rank Jacobians everywhere, i.e. be a local diffeomorphism, \label{prop:3}
  \item intervene rarely, and \label{prop:4}
  \item compute quickly.  \label{prop:5}
\end{properties}
\par
Subsequently, we present two safeguards that offer different trade-offs between the desired properties. We summarise the properties of the safeguards in \Cref{tab:safeguards}.
\begin{table*}
  \centering
  \caption{Properties of the unaltered safeguards.}
  \label{tab:safeguards}
  \begin{tabular}{l r r}
    \toprule
                                           & Boundary Projection                               & Ray Mask                                                \\
    \midrule
    \Cref{prop:1} Safe                     & \cmark                                            & \cmark                                                  \\
    \Cref{prop:2} Subdifferentiable        & \cmark                                            & \cmark                                                  \\
    \Cref{prop:3}                          &                                                   &                                                         \\
    \quad Class $C^1$                      & almost everywhere                                 & almost everywhere                                       \\
    \quad Jacobian rank                    & $d - 1$                                           & \cmark($d$)                                             \\
    \Cref{prop:4} Interventions            & $\forall a \in \mathcal{A}\setminus\mathcal{A}_s$ & $\forall a \in \mathcal{A}$                             \\

    \Cref{prop:5} Computational complexity &                                                   &                                                         \\
    \quad Specified $\mathcal{A}_{s}$
                                           & 1 Quadratic Program                               & 1 Linear Program                                        \\
    \quad Derived $\mathcal{A}_{s}$
                                           & 1 Quadratic Program                               & (1 Conic $\vee$ 1 Quadratic)  $\wedge$ 1 Linear Program \\
    \bottomrule
  \end{tabular}
\end{table*}
\subsection{BOUNDARY PROJECTION}
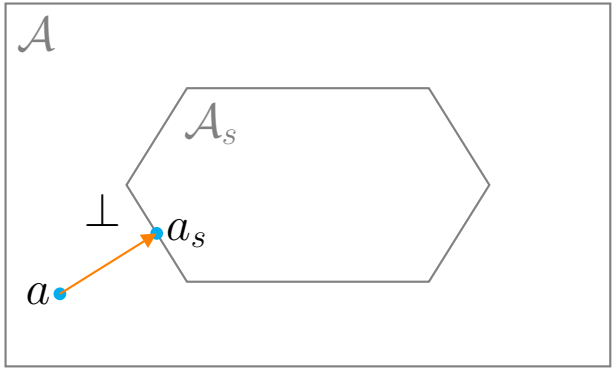
\begin{figure}
  \centering
  \begin{tikzpicture}[scale=1.6, every node/.style={font=\Huge}]

\draw[gray, thick] (-2.5,-1.5) rectangle (2.5,1.5);
\node[text=gray] at (-2.25,1.25) {$\mathcal{A}$};

\draw[gray, thick] 
    (-1,0.8) -- (1,0.8) -- (1.5,0) -- (1,-0.8) -- (-1,-0.8) -- (-1.5,0) -- cycle;
\node[text=gray] at (-0.8,0.5) {$\mathcal{A}_s$};

\coordinate (as) at (-1.25, -0.4);
\coordinate (a) at (-2.05, -0.9);

\node at (-1.7, -0.2) {$\perp$};

\fill[cyan] (a) circle (1.5pt);
\node[left] at (a) {$a$};

\fill[cyan] (as) circle (1.5pt);
\node[right] at (as) {$a_s$};

\draw[-{Triangle[length=2mm,width=2mm]},orange,thick] (a) -- (as);

\end{tikzpicture}
  \caption{Boundary projection maps unsafe actions to the boundary of the safe action set by determining the closest safe action.}
  \label{fig:border_projection}
\end{figure}

The boundary projection safeguard, proposed in \cite{BoundaryProjection}, maps any action to the closest safe action. By definition, it therefore only affects unsafe actions, which are mapped to the nearest boundary point in the safe action set. In Euclidean space, this corresponds to an orthogonal projection to the boundary of the safe action set. We show an exemplary mapping with boundary projection from an unsafe action $a$ to a safe action $a_s$ in \Cref{fig:border_projection}.
The safeguard $g_{\text{BP}}(a)$ provides the safe action by solving
\begin{subequations}
  \label{eq:boundary_projection}
  \begin{alignat}{2}
     & \!\min_{a_{s}}    & \norm{a-a_s}^2                                     \\
     & \text{subject to} & a_s \in \mathcal{A}_s \label{constraint:bp_1} \, .
  \end{alignat}
\end{subequations}

\subsubsection{PROPERTIES}
The optimisation problem is always solvable given \Cref{proposition:existence}, and \Cref{constraint:bp_1} ensures satisfaction of \Cref{prop:1}. The implicit function theorem \cite[Theorem 3.3.1]{ImplicitFunctionTheorem} provides the Jacobian of the solution mapping to satisfy \Cref{prop:2}.
\par
The distance between the initial and mapped action decreases smoothly with the distance from the unsafe action to the boundary until it is zero for safe actions. However, the mapping location can change abruptly between unsafe actions on different sides of the edges of the safe set. This leads to a jump in the gradient, such that it is only $C^1$ almost everywhere. We employ any element from the Clarke subdifferentiable \cite{Clarke} at non-differentiable points, ensuring our approach remains well-defined. In addition, all actions along the ray starting at a boundary point in the direction of the outward normal are mapped to that boundary point. Formally, any unsafe action $a_u$ that can be written as
\begin{equation}
  \forall t>0 \,: a_u = a_{s, \partial\mathcal{A}_s} + t \cdot v
\end{equation}
with the safe action on the boundary $a_{s, \partial\mathcal{A}_s} \in \partial \mathcal{A}_s$ and $v$ any outward normal vector at $a_{s, \partial\mathcal{A}_s}$, is mapped by \Cref{eq:boundary_projection} to $a_{s, \partial\mathcal{A}_s} $. Therefore, the safeguard cannot propagate gradients in the mapping direction, such that
\begin{equation}
  \label{eq:boundary_projection_gradient}
  \left(\frac{\partial r}{\partial a_s}\frac{\partial a_s}{\partial a}\right)v = 0 \, ,
\end{equation}
which is especially problematic for gradients parallel to $v$. In such a case, boundary projection eliminates the gradient, keeping the optimisation stuck indefinitely.
\begin{theorem}
  \label{lem:boundary_projection}
  Let \(\mathcal{A}_s\) be the zonotope $\left<c_{\mathcal{A}_s},G_{\mathcal{A}_s}\right>$ with generator matrix $G_{\mathcal{A}_s} \in \mathbb{R}^{d\times n}$, such that strict complementary slackness holds for \Cref{eq:boundary_projection}, and $g_{\text{BP}}$ be differentiable. Then the rank of the Jacobian of \Cref{eq:boundary_projection} is
  \begin{equation}
    \text{rank}\left(\frac{\partial a_s}{\partial a}\right) = \begin{cases}
      d   & \text{if } a \in \mathcal{A}_s \\
      < d & \text{else} \, .
    \end{cases}
  \end{equation}
\end{theorem}
We obtain a proof by differentiating through the Karush-Kuhn-Tucker (KKT) conditions, which we provide in \Cref{app:jacobian}. Consequently, boundary projection does not satisfy \Cref{prop:3}.
\par
Boundary projection only intervenes for unsafe actions and therefore adheres to \Cref{prop:4}. If the safe action set is specified, \Cref{eq:point_containment} is the containment \Cref{constraint:bp_1}. Otherwise, we use the constraint $\mathcal{S}_{i+1}(a_i, s_i)\subseteq \mathcal{S}_s$, which is tightened by \Cref{eq:set_containment} and convex for a linearised transition function, as \Cref{eq:set_containment} is a linear constraint. Both yield quadratic programs, which compute quickly.

\subsubsection{MODIFICATIONS}
To regain a gradient in the mapping direction and compensate for the resulting rank-deficient Jacobian, which violates \Cref{prop:3}, we augment the policy loss function $l_r(a_s, s)$ with a regularisation term \cite[Eq. 16]{BoundaryProjectionLoss}
\begin{equation}
  l(a, s, a_s) = l_r(a_s, s) + c_d \norm{a_s - a}^2 \, .
  \label{eq:L2loss}
\end{equation}
As a result, the corresponding gradient
\begin{equation}
  \frac{\partial l}{\partial a} = \frac{\partial l_r}{\partial a} + 2c_d(a_s-a)^T\left(\frac{\partial a_s}{\partial a} - I\right)
\end{equation} points along the projection direction $a_s -a$. The coefficient $c_d$ scales the regularisation to remain small relative to the original loss $l_r(a_s, s)$, yet large enough to produce a meaningful gradient in the mapping direction. In addition to gradient augmentation, the regularisation encourages the policy to favour safe actions from the start, which is desirable as stated in \Cref{prop:4}.

\subsection{RAY MASK}
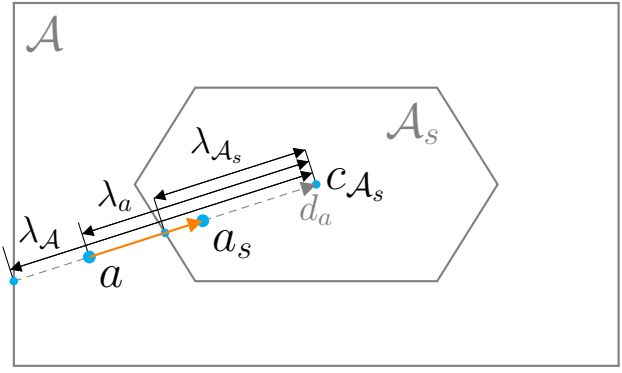
\begin{figure}
  \centering
  \begin{tikzpicture}[scale=1.6, every node/.style={font=\Huge}]

\draw[gray, thick] (-2.5,-1.5) rectangle (2.5,1.5);
\node[text=gray] at (-2.25,1.25) {$\mathcal{A}$};

\draw[gray, thick] 
    (-1,0.8) -- (1,0.8) -- (1.5,0) -- (1,-0.8) -- (-1,-0.8) -- (-1.5,0) -- cycle;
\node[text=gray] at (0.8,0.5) {$\mathcal{A}_s$};

\coordinate (n_da) at (-0.30478, +0.952420);
\coordinate (cs) at (0.0, 0.0);
\coordinate (bas) at (-1.25, -0.4);
\coordinate (as) at (-0.93749, -0.3);
\coordinate (a) at (-1.875, -0.6);
\coordinate (ba) at (-2.5, -0.8);

\draw[black, thin] (cs) -- ($0.3*(n_da)$);
\draw[black, thin] (bas) -- ($(bas)+0.3*(n_da)$);
\draw[black, thin] (a) -- ($(a)+0.3*(n_da)$);
\draw[black, thin] (ba) -- ($(ba)+0.3*(n_da)$);

\draw[-{Triangle[length=1.5mm,width=1.5mm]}] ($0.1*(n_da)$) -- ($(ba)+0.1*(n_da)$);
\draw[-{Triangle[length=1.5mm,width=1.5mm]}] ($(ba)+0.1*(n_da)$) -- ($0.1*(n_da)$);
\node[above, font=\Large] at ($0.5*(ba)+0.5*(a) + (-0.1, 0.1)$) {$\lambda_{\mathcal{A}}$};

\draw[-{Triangle[length=1.5mm,width=1.5mm]}] ($0.2*(n_da)$) -- ($(a)+0.2*(n_da)$);
\draw[-{Triangle[length=1.5mm,width=1.5mm]}] ($(a)+0.2*(n_da)$) -- ($0.2*(n_da)$);
\node[above, font=\Large] at ($0.5*(bas)+0.5*(a) + (-0.1, 0.2)$) {$\lambda_a$};

\draw[-{Triangle[length=1.5mm,width=1.5mm]}] ($0.3*(n_da)$) -- ($(bas)+0.3*(n_da)$);
\draw[-{Triangle[length=1.5mm,width=1.5mm]}] ($(bas)+0.3*(n_da)$) -- ($0.3*(n_da)$);
\node[above, font=\Large] at ($0.5*(as)+0.5*(cs) + (-0.35, 0.25)$) {$\lambda_{\mathcal{A}_s}$};

\fill[cyan] (cs) circle (1.pt);
\node[right] at (cs) {$c_{\mathcal{A}_s}$};
\draw[-{Triangle[length=2mm,width=2mm]}, gray, thin, dash pattern=on 3pt off 2pt] (ba) -- (cs);
\node[below, text=gray, font=\Large] at (cs) {$d_a$};

\fill[cyan] (bas) circle (1.pt);
\fill[cyan] (ba) circle (1.pt);

\fill[cyan] (a) circle (1.5pt);
\node[below right] at (a) {$a$};

\fill[cyan] (as) circle (1.5pt);
\node[below right] at (as) {$a_s$};

\draw[-{Triangle[length=2mm,width=2mm]},orange,thick] (a) -- (as);

\end{tikzpicture}
  \caption{The ray mask maps actions towards the safe centre $c_{\mathcal{A}_s}$ in proportion to the safety domain length $\lambda_{\mathcal{A}_s}$, the feasible domain length $\lambda_\mathcal{A}$, and the distance from the action to the safe centre $\lambda_a$.}
  \label{fig:ray_masking}
\end{figure}
The ray mask \cite{RayMasking} maps every action radially towards the centre of the safe action set $c_{\mathcal{A}_s}$, as shown in \Cref{fig:ray_masking}. Using the unit vector $d_a = \frac{a-c_{\mathcal{A}_s}}{\norm{a - c_{\mathcal{A}_s}}}$, we define the distances from the safe action set centre to the initial action and the boundaries of the safe and feasible action sets as
\begin{align}
  \lambda_a               & = \norm{a - c_{\mathcal{A}_s}} \label{eq:lambda_a}                                                               \\
  \lambda_{\mathcal{A}_s} & = \max\{ \lambda \geq 0 \mid c_{\mathcal{A}_s} + \lambda d_a \in \mathcal{A}_s\} \label{eq:lambda_a_safe}        \\
  \lambda_{\mathcal{A}}   & = \max\{ \lambda \geq 0 \mid c_{\mathcal{A}_s} + \lambda d_a \in \mathcal{A}\} \, . \label{eq:lambda_a_feasible}
\end{align}
We introduce the generalised ray mask as
\begin{equation}
  g_{\text{RM}}(a) = \begin{cases}
    c_{\mathcal{A}_s}                                    & \text{if } \norm{a - c_{\mathcal{A}_s}} < \epsilon \\
    c_{\mathcal{A}_s} + \omega\lambda_{\mathcal{A}_s}d_a & \text{else,}
  \end{cases} \label{eq:generalised_rm}
\end{equation}
where $\varepsilon \ll 1$ ensures numerical stability, and the mapping function, whose arguments were omitted for clarity, $\omega(\lambda_a, \lambda_{\mathcal{A}_s}, \lambda_{\mathcal{A}})$ is characterised by:
\begin{equation}
  \omega(\lambda_a, \lambda_{\mathcal{A}_s}, \lambda_{\mathcal{A}}) \colon (0, \lambda_\mathcal{A}]^2 \times \mathbb{R}_{>0} \mapsto (0, 1] \label{eq:mapping_func}
\end{equation}
\begin{equation}
  \frac{\partial \omega(\lambda_a, \lambda_{\mathcal{A}_s}, \lambda_{\mathcal{A}})}{\partial \lambda_a} > 0 \, , \label{eq:mapping_deriv}
\end{equation}
which ensures a safe, convex mapping. The linear ray mask introduced in \cite[Eq. 6]{RayMasking} is obtained by setting $w_{\text{lin}}(\lambda_a, \lambda_{\mathcal{A}_s}, \lambda_{\mathcal{A}}) = \frac{\lambda_a}{\lambda_\mathcal{A}}$.
\par
In addition to the constraints introduced in the problem statement in \Cref{sec:problem_statement}, ray masking requires a star-shaped \cite[Def. 5.2.9]{papadopoulos2005metric} safe action set to ensure that the safe centre and the line segment from the safe boundary point to the safe centre lie within the set. All convex sets, including zonotopes, are star-shaped. While a derived safe action set is not necessarily convex, it is convex for a linearised transition function.
\begin{theorem}
  \label{lem:ray_map}
  Let $\mathcal{S}_s$ be a zonotope and $\mathcal{S}_{i+1}(a_i, s_i)$ be the next state set as in \Cref{eq:next_state_set}. Then, $\mathcal{A}_s$ in \Cref{eq:safe_action_set} is convex.
\end{theorem}
\begin{IEEEproof}
  We start by inserting \Cref{eq:next_state_set} into \Cref{eq:safe_action_set}
  \begin{equation}
    \mathcal{A}_s(s_i) = \left\{a_i \in \mathcal{A} \mid M a_i \oplus \left<c + c_\mathcal{W}, G_{\mathcal{W}}\right> \subseteq \mathcal{S}_{s} \right\} \, .
  \end{equation}
  The safe action set is convex if and only if the set of all translations $\mathcal{T} = \left\{t \mid t \oplus \left<c + c_\mathcal{W}, G_{\mathcal{W}}\right> \subseteq \mathcal{S}_s\right\}$ is convex. This is the definition of a Minkowski difference, which is convexity preserving \cite[Theorem 2.1]{mindiff}.
\end{IEEEproof}
\subsubsection{COMPUTATION}
\begin{figure}
  \centering
  \begin{tikzpicture}[scale=0.8, every node/.style={font=\Large}]
    \begin{scope}[shift={(-2.7,0)}] 
      \draw[gray, thick] (-2.5,-1.5) rectangle (2.5,1.5);
      \node[text=gray] at (-2.25,1.25) {$\mathcal{A}$};

      \draw[gray, thick] 
          (-1.2,1.0) -- (1.2,1.0) -- (1.7,0) -- (1.2,-1.0) -- (-1.2,-1.0) -- (-1.7,0) -- cycle;

      \coordinate (cs) at (0.0, 0.0);
      \fill[cyan] (cs) circle (1.5pt);
      \node[below] at (cs) {$c_{\mathcal{A}_s}$};

      \draw[green, thick] 
          (-0.5, 1.0) -- (0.5, 1.0) -- (1.4, 0.6) -- (1.4, 0.4) -- (1.4, -0.4) -- (1.4, -0.6) -- 
          (0.5, -1.0) -- (-0.5, -1.0) -- (-1.4, -0.6) -- (-1.4, -0.4) -- (-1.4, 0.4) -- (-1.4, 0.6) -- cycle;

      \node at (0.2, 0.6) {$\mathcal{Z}_{\mathcal{A}_s}$};
      \node[text=gray] at (-1.0,0.7) {$\mathcal{A}_s$};
    \end{scope}

    \begin{scope}[shift={(2.7,0)}] 
      \draw[gray, thick] (-2.5,-1.5) rectangle (2.5,1.5);
      \node[text=gray] at (-2.25,1.25) {$\mathcal{A}$};

      \draw[gray, thick] 
          (-1.2,1.0) -- (1.2,1.0) -- (1.7,0) -- (1.2,-1.0) -- (-1.2,-1.0) -- (-1.7,0) -- cycle;

      \coordinate (a) at (-2.03, -0.58);
      \coordinate (as) at (-1.53, -0.33);
      \coordinate (bas) at (1.12, 0.995);
      \coordinate (n) at (-1.575, 3.15);

      \draw[black, thick] (bas) -- ($(bas) + 0.05*(n)$);
      \draw[black, thick] (as) -- ($(as) + 0.05*(n)$);
      \draw[-{Triangle[length=1.5mm,width=1.5mm]}] ($(bas) + 0.05*(n)$) -- ($(as) + 0.05*(n)$);
      \draw[-{Triangle[length=1.5mm,width=1.5mm]}] ($(as) + 0.05*(n)$)-- ($(bas) + 0.05*(n)$) ;
      \node[above] at ($0.5*(bas)+0.5*(as) + (-0.5, 0.0)$) {$\lambda_\perp$};

      \draw[-{Triangle[length=2mm,width=2mm]}, gray, thin, dash pattern=on 3pt off 2pt]  (a) -- (bas);

      \fill[cyan] (a) circle (1.5pt);
      \node[below right] at (a) {$a$};

      \fill[cyan] (as) circle (1.5pt);
      \node[below right] at (as) {$a_{s,\text{BP}}$};

      \fill[cyan] ($0.5*(bas)+0.5*(as)$) circle (1.5pt);
      \node[below right] at ($0.5*(bas)+0.5*(as)$) {$c_{\mathcal{A}_s}$};

      \fill[cyan] (bas) circle (1.5pt);
      \node[text=gray] at (1.1, 0.5) {$d_\perp$};

      \node[text=gray] at (0.8, -0.7) {$\mathcal{A}_s$};
    \end{scope}
  \end{tikzpicture}
  \caption{Zonotopic approximation of the safe centre $c_{\mathcal{A}_s}$ by expanding a contained zonotope $\mathcal{Z}_{\mathcal{A}_s}$ (left) and orthogonal approximation by piercing the safe action set $\mathcal{A}_s$ orthogonal to the boundary and taking the midpoint between both boundary points $a_{s,\text{BP}}$ and $a_{s,\text{BP}} + \lambda_\perp d_\perp$ (right).}
  \label{fig:approximations}
\end{figure}
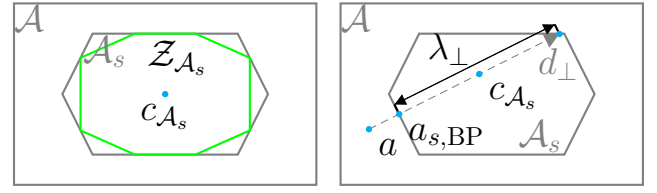

The distance to the safe action set $\lambda_{\mathcal{A}_s}$ can be computed by
\begin{subequations}
  \label{cvx:boundary}
  \begin{alignat}{2}
     & \!\max_{\lambda_{\mathcal{A}_s}} & \quad & \lambda_{\mathcal{A}_s}                                                                           \\
     & \text{subject to}                & \quad & c_{\mathcal{A}_s} + \lambda_{\mathcal{A}_s} d_a \in \mathcal{A}_s \label{constraint:boundary}\, .
  \end{alignat}
\end{subequations}
The distance to the feasible action set $\lambda_\mathcal{A}$ can be computed equivalently if it is a zonotope. For an axis-aligned box, the computation is possible in closed-form \cite{aabb}. For a specified safe action zonotope, the safe centre is defined as the centre of the zonotope. The safe centre is not readily available for a derived safe action set, as in \Cref{eq:safe_action_set}. We present two approaches to approximate it: orthogonal and zonotopic approximation, which are visualised in \Cref{fig:approximations}.
\par
The zonotopic approach directly approximates the safe action set by maximising the generator lengths of a zonotope while maintaining containment. The under-approximated zonotope $\mathcal{Z}_{\mathcal{A}_s}$ is the solution to
\begin{subequations}
  \label{cvx:zonotope_expansion}
  \begin{alignat}{2}
     & \!\max_{c_{\mathcal{A}_s}, l_s} & \quad & \prod_{i=1}^{n}l_{s, i} \label{eq:expansion_objective}                                  \\
     & \text{subject to}               & \quad & \mathcal{Z}_{\mathcal{A}_s} = \left<c_{\mathcal{A}_s}, G_{\mathcal{A}_s} (l_s)_D\right> \\
     &                                 &       & \mathcal{Z}_{\mathcal{A}_s} \subseteq \mathcal{A}                                       \\
     &                                 &       & \mathcal{S}_{i+1}(\mathcal{Z}_{\mathcal{A}_s}, s_i) \subseteq \mathcal{S}_s
  \end{alignat}
\end{subequations}
with $n$ generator directions $G_{\mathcal{A}_s}$ sampled uniformly from a $d$-dimensional sphere $\mathbb{S}^d$ and where we denote the diagonalisation of a vector by the subscript $D$. Generally, the number of generators should be in the order of magnitude of the action dimension to provide a good approximation. However, $n$ should also not be too large, since we employ the volume computation of a box as a computationally cheaper proxy for the volume of a zonotope \cite{RayMasking}, which assumes orthogonal generators. This assumption is violated for $n>d$. Therefore, the objective \Cref{eq:expansion_objective} favours spherical zonotopes over elongated ones, which is not necessarily volume-maximising if the proper safe action set is elongated.
\par 
The orthogonal approximation computes the required safe centre $c_{\mathcal{A}_s}$ and distances $\lambda_a$, $\lambda_\mathcal{A}$, and $\lambda_{\mathcal{A}_s}$ without the expensive approximation of the safe action set. Instead, it pierces the safe action set orthogonal to the boundary and assumes the midpoint between the entry and exit point as the safe centre. The orthogonal starting point and direction is determined by \Cref{eq:boundary_projection}, which yields $a_{s,\text{BP}}$ and $d_{\perp} = \frac{a_{s,\text{BP}} - a}{\norm{a_{s,\text{BP}} - a}}$. Next, we reuse \Cref{cvx:boundary} as
\begin{subequations}
  \begin{alignat}{2}
     & \!\max_{\lambda_{\perp}} & \quad & \lambda_{\perp}                                                   \\
     & \text{subject to}        & \quad & a_{s, \text{BP}} + \lambda_{\perp} d_\perp \in \mathcal{A}_s \, .
  \end{alignat}
\end{subequations}
We utilise the midpoint between $a_{s,\text{BP}}$ and $a_{s,\text{BP}} + \lambda_\perp d_\perp$ as the safe centre
\begin{equation}
  c_{\mathcal{A}_s} = a_{s, \text{BP}} + \frac{\lambda_\perp}{2}d_\perp \, .
\end{equation}
Since \Cref{eq:boundary_projection} will only yield a different action for unsafe actions, the orthogonal approximation technique is restricted to those actions.

\subsubsection{PROPERTIES}
The generalised ray mask satisfies \Cref{prop:1}, since its codomain is the safe action set. To illuminate this fact, we remark that \Cref{eq:generalised_rm} can be examined in one dimension -- the direction along the ray $d_a$ -- without loss of generality. The action along this ray is bounded between $c_{\mathcal{A}_s}$, which maps to $g_{\text{RM}}(c_{\mathcal{A}_s}) = c_{\mathcal{A}_s} \in \mathcal{A}_s$, and $c_{\mathcal{A}_S} +\lambda_\mathcal{A} d_a$, which maps to $g_{\text{RM}}(c_{\mathcal{A}_S} +\lambda_\mathcal{A} d_a) = c_{\mathcal{A}_s}+\lambda_{\mathcal{A}_s}d_a \in \mathcal{A}_s$. Gradients to obtain \Cref{prop:2} are available from backpropagating through \Cref{eq:generalised_rm}.
\par
Regarding smoothness, the ray mask safeguard is of class $C^1$ almost everywhere, except for the safe set edges and for the $\epsilon$-sphere around the safe action set centre $\mathcal{A}_\epsilon = \{a \in \mathcal{A}_s \mid \norm{a - c_{\mathcal{A}_s}} = \epsilon\}$. As in boundary projection, we employ any element from the Clarke subdifferentiable \cite{Clarke} at these edges. The Jacobian of a ray mask has full rank, wherever $g_\text{RM}$ is differentiable. Consequently, a ray mask satisfies \Cref{prop:3} almost everywhere.
\begin{theorem}
  \label{lem:ray_map_gradient}
  Let \(\mathcal{A}_s\) be convex, $g_\text{RM}$ differentiable and $\norm{a - c_{\mathcal{A}_s}} > \epsilon$. Then, the Jacobian of any ray mask as in \Cref{eq:generalised_rm} has full rank.
\end{theorem}
We present the proof in \Cref{app:ray_mask_jacobian_rank}. While the ray mask propagates gradients in the mapping direction, they are still diminished for the linear mapping. This reduction is particularly obvious for the linear mapping function in the scenario where the feasible and safe action set are spheres with coinciding centres and radii $r_{\mathcal{A}} > r_{\mathcal{A}_s}$, and the coordinate system is already spherical and centred. In this scenario, the Jacobian in \Cref{eq:ray_gradient} reduces to
\begin{equation}
  \frac{\partial a_s}{\partial a} = \begin{bmatrix}
    \frac{r_{\mathcal{A}_s}}{r_{\mathcal{A}}} & \textbf{0} \\
    \textbf{0}                                & I
  \end{bmatrix} \, ,
\end{equation}
which has a trivial eigenspace, as the Jacobian is diagonal. Consequently, the upstream gradient is only modified in the mapping direction by the factor $\frac{r_{\mathcal{A}_s}}{r_{\mathcal{A}}} < 1$. \\
Contrary to the boundary projection safeguard, the ray mask applies to all actions, including safe actions. Moreover, the linear mapping distance decreases only linearly with the distance to the safe centre, as the partial derivative is constant in $\lambda_a$:
\begin{equation}
  \frac{\partial \omega_{\text{lin}}}{\partial \lambda_a} = \frac{1}{\lambda_\mathcal{A}} \, .
\end{equation}This means that safe actions far from the centre are also substantially altered, therefore \Cref{prop:4} is not firmly adhered to.
\par 
In regard to \Cref{prop:5} and computational complexity, the actual application of the ray mask in \Cref{eq:generalised_rm} is negligible, as it is a closed-form expression. However, computing the safe boundary \Cref{cvx:boundary} is a linear program, since \Cref{constraint:boundary} has to be considered through \Cref{eq:point_containment} or \Cref{eq:set_containment}, depending on the availability of the safe action set. A specified safe action set provides the safe centre. However, for derived safe action sets, as in \Cref{eq:safe_action_set}, the approximations can be costly. For linearised dynamics, the zonotopic approach is a conic program, while the orthogonal approximation requires the solution of one quadratic program.

\subsubsection{MODIFICATIONS}
We propose three possible modifications to the linear ray mask to improve its learning properties. First, we can increase the gradient in the mapping direction with the same regularisation term as in \Cref{eq:L2loss} to compensate for the diminished gradient and nudge towards safety.
\par
Second, we can replace the Jacobian with an identity matrix for faster computation and unimpeded gradient propagation, which we denote passthrough. This modification retains the correct gradient directions if the reward-maximising action is safe. However, whether the reward-maximising action is safe is generally unknown and depends on the environment. For unsafe reward-maximising actions, the point of convergence of the policy optimisation would be the safe boundary point on the line $\overline{c_{\mathcal{A}_s}a_{r_{max}}}$, which is no longer optimal.
\par
\begin{figure*}
  \centering
  \includegraphics[width=\linewidth]{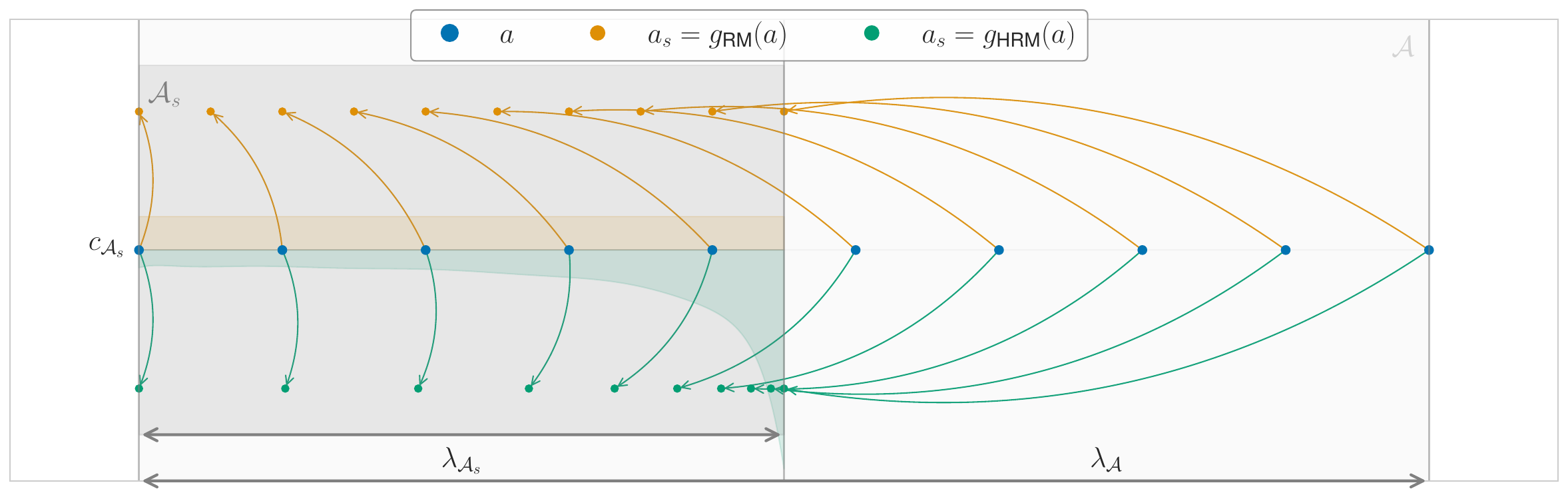}
  \caption{One-dimensional illustration of the linear ray mask (RM) and the hyperbolic ray mask (HRM). Arrows show mappings for exemplary unsafe actions to the corresponding safeguarded actions. Shaded regions indicate the distribution of safe actions. The linear ray mask maps $\mathcal{A}$ linearly onto $\mathcal{A}_s$, whereas the hyperbolic ray mask maps it exponentially, such that unsafe actions are projected closely to the safe boundary and safe actions are perturbed minimally.}
  \label{fig:mappings}
\end{figure*}

Finally, we propose a hyperbolic mapping function to limit the perturbation of safe actions.
We visually compare both mappings in \Cref{fig:mappings}.
The hyperbolic mapping is defined as
\begin{equation}
  \omega_{\text{tanh}}(\lambda_a, \lambda_{\mathcal{A}_s}, \lambda_\mathcal{A}) = \frac{\tanh{\frac{\lambda_a}{\lambda_{\mathcal{A}_s}}}}{\tanh{\frac{\lambda_\mathcal{A}}{\lambda_{\mathcal{A}_s}}}} \, ,
\end{equation}
which maps unsafe actions close to the boundary and safe actions close to themselves, since $\omega_{\text{tanh}}(\lambda_a  > \lambda_{\mathcal{A}_s}) \approx \lambda_{\mathcal{A}_s}$ and $\omega_{\text{tanh}}(\lambda_a  < \lambda_{\mathcal{A}_s}) \approx \frac{\lambda_a}{\lambda_{\mathcal{A}_s}}$. It is a valid mapping, as defined in \Cref{eq:mapping_func} and \Cref{eq:mapping_deriv}, as $w_{\text{tanh}}(\lambda_a=\lambda_\mathcal{A})=1$, $w_{\text{tanh}}(\lambda_a=0) = 0$, and
\begin{equation}
  \frac{\partial \omega_{\text{tanh}}}{\partial \lambda_a} = \frac{1-\tanh^2{\frac{\lambda_a}{\lambda_{\mathcal{A}_s}}}}{\lambda_{\mathcal{A}_s}\tanh{\frac{\lambda_\mathcal{A}}{\lambda_{\mathcal{A}_s}}}} > 0 \, ,
\end{equation}
since $\lambda_{\mathcal{A}_s}, \lambda_{\mathcal{A}} >0$ and $\tanh^2: \mathbb{R} \mapsto [0, 1)$. The hyperbolic map maintains the idea of a radial mapping towards the safe centre, while its mapping behaviour is similar to boundary projection in terms of mapping distance; it nonetheless provides a full Jacobian and a smooth mapping for unsafe and safe actions. Due to its similarity to boundary projection, it also has reduced gradients in the ray direction and benefits from a regularisation term.

\section{NUMERICAL EXPERIMENTS}
This section tests our two main hypotheses:
\begin{hypotheses}
  \item  Under safeguarding, analytic gradient-based reinforcement learning achieves higher evaluation returns from fewer environment interactions than sampling-based reinforcement learning. \label{hyp:1}
  \item Enabling our modified safeguards for analytic gradient-based reinforcement learning during training leads to similar or higher return policies than unsafe training, given the same number of environment interactions. \label{hyp:2}
\end{hypotheses}
The following subsections introduce our experimental setup, discuss the main hypotheses, and provide additional insights.

\subsection{SETUP}
We conducted all experiments using ten different random seeds. Hyperparameters were tuned exclusively for non-safeguarded training and carried over unchanged to the safeguarded experiments~\cite{Optuna}. We assessed the quality of the final policy by calculating the return \mbox{(Return)} achieved over a representative evaluation set. We tracked the number of steps until convergence \mbox{(\# Steps)}, defined as reaching within 5\% of the return of the final policy. Further, we report the mean and a 95\% confidence interval computed using bootstrapping for both the return and number of steps. Lastly, we report the number of runs that did not converge within the maximum allowed number of environment interactions \mbox{(\# Stuck)}. We excluded non-convergent runs from the return and step calculations to ensure clarity.
\par
In our numerical experiments, we vary all three components of the policy optimisation: learning algorithm, safeguarding, and environment.
\subsubsection{LEARNING ALGORITHMS}
We choose the first-order reinforcement learning algorithm SHAC \cite{SHAC} over its successor adaptive-horizon actor-critic \cite{AHAC} due to its maturity, stable convergence, and the lack of stiff dynamics in our tasks. We compare it with two well-established sampling-based reinforcement learning algorithms: on-policy proximal policy optimisation (PPO) \cite{PPO} and off-policy soft actor-critic (SAC) \cite{SAC}.
\par
Replacing unsafe actions with safe actions inside the policy poses problems for stochastic policies, which rely on the probabilities of the actions. We therefore implement safeguarding as a post-processing step to the policy output without explicitly informing the sampling-based learning processes. This requires them to learn the dynamics associated with the safeguarded environment.
\subsubsection{SAFEGUARDS}
We evaluate the base versions of the boundary projection and ray mask as safeguards, where we approximate the safe centre using the zonotopic approach. We also assess all modifications to the safeguards individually, as well as the combination of regularisation and the hyperbolic ray mask.
\subsubsection{ENVIRONMENTS}
We study three environments, which are detailed in \Cref{app:envs}. The first two are balancing tasks for a pendulum and a quadrotor, where we minimise the distance to an equilibrium position. The safety constraints comprise both action and state constraints that limit, for example, angles and angular velocities. To guarantee constraint satisfaction at all times, we use robust control invariant sets \cite{RCI-Sets} as time-invariant safe state sets. The third environment features an energy management system for a battery and a heat pump aimed at minimising the electricity cost of a household while maintaining a comfortable indoor temperature. Both the state of charge of the battery and the room temperature have limits that must be enforced at all times. When considering the full action range, we achieve this by computing a safe state set that ensures that the system can be steered back into the feasible set within one time step.
\par
We build our differentiable simulations according to the gymnasium framework \cite{Gymnasium} and differentiate through the dynamics using PyTorch's auto-differentiation engine \cite{Pytorch}. We formulate the convex optimisation problems with CVXPY \cite{CVXPY, CVXPY2} and backpropagate through them with CVXPYLayers \cite{CVXPY-Layers}.

\subsection{EVALUATION OF LEARNING ALGORITHMS}
In this subsection, we evaluate \Cref{hyp:1} by comparing the sampling-based reinforcement learning algorithms PPO and SAC with the analytic gradient-based algorithm SHAC.
\par
We first compared the learning algorithms in unsafe training to establish a baseline. The key metrics are listed in \Cref{tab:unsafe} and the learning curves in \Cref{app:unsafe_results}. SHAC converged to the best policies in the pendulum and quadrotor tasks, where it was the only algorithm to balance the quadrotor consistently with minimal effort. However, it performed substantially worse on the energy system task. PPO performed best in energy systems but worst in the pendulum and quadrotor environments.
\par
We attribute the performance degradation of SHAC in the energy system task to the high degree of noise of the environment. Environmental noise likely disrupts the computation of meaningful analytic gradients, and the smooth surrogate critic employed by SHAC may poorly approximate the true reward landscape. In contrast, PPO does not rely on a smooth reward approximation and benefits from the large number of simulation interactions available in the energy system task. Nevertheless, both PPO and SHAC had runs that failed to learn a meaningful policy in the energy system environment. This was also the case for one SHAC run in the pendulum environment.

\begin{table*}
  \centering
  \caption{Comparison of learning algorithms in unsafe training.}
  \label{tab:unsafe}
  \begin{tabular}{l l r r r r r}
    \toprule
    \multirow{2}{*}{Environment} & \multirow{2}{*}{Algorithm} & \multicolumn{2}{c}{\# Step} & \multicolumn{2}{c}{Return}           & \multirow{2}{*}{\# Stuck}                                                  \\
    \cmidrule(l){3-4}\cmidrule(l){5-6}
                                 &                            & Mean                        & 95\% CI                              & Mean                      & 95\% CI                               &        \\
    \midrule
    \multirow{3}{*}{Pendulum}

                                 & SHAC                       & 12800                       & [\phantom{0}10808, \phantom{0}15360] & \textbf{-8}               & [\phantom{00000}-8, \phantom{0000}-8] & 1 / 10 \\
                                 & SAC                        & 8513                        & [\phantom{00}6260, \phantom{0}10767] & -14                       & [\phantom{0000}-15, \phantom{000}-11] & 0 / 10 \\
                                 & PPO                        & 81600                       & [\phantom{0}81600, \phantom{0}81600] & -596                      & [\phantom{00}-1174, \phantom{-00}300] & 0 / 10 \\

    \midrule
    \multirow{3}{*}{Quadrotor}

                                 & SHAC                       & 20364                       & [\phantom{0}11558, \phantom{0}28070] & \textbf{-157}             & [\phantom{000}-169, \phantom{00}-140] & 0 / 10 \\
                                 & SAC                        & 80628                       & [\phantom{0}60096, 109174]           & -1046                     & [\phantom{00}-1855, \phantom{-00}170] & 0 / 10 \\
                                 & PPO                        & 80640                       & [\phantom{0}42560, 116480]           & -1710                     & [\phantom{00}-2128, \phantom{0}-1267] & 0 / 10 \\

    \midrule
    \multirow{3}{*}{Energy System}

                                 & SHAC                       & 259600                      & [\phantom{0}89393, 400400]           & -114164                   & [-146048, -79760]                     & 1 / 10 \\
                                 & SAC                        & 674999                      & [554974, 781998]                     & -5225                     & [\phantom{00}-9424, \phantom{-00}353] & 0 / 10 \\
                                 & PPO                        & 748800                      & [645120, 861120]                     & \textbf{-2739}            & [\phantom{00}-4598, \phantom{00}-167] & 2 / 10 \\
    \bottomrule
  \end{tabular}
\end{table*}
\par
After establishing the baselines in unsafe training, we proceed to testing the hypothesis by comparing the learning algorithms in safeguarded training. We show the key metrics in \Cref{tab:safe} and the learning curves in \Cref{app:results_bp} and \Cref{app:results_rm}. We obtained results similar to unsafe training, as SHAC converged to the best policies in the balancing tasks, whereas the sampling-based methods outperformed SHAC on the energy system. However, SHAC performed substantially better in safeguarded training in the energy system task.
\par
SAC in safeguarded training showed volatile behaviour in balancing scenarios. For example, in the pendulum task, SAC initially reached near-optimal performance within the first evaluation but later diverged. In the quadrotor environment, SAC learned ineffectively until the buffer reset roughly twice, at which point a jump in performance was consistently visible. Under uninformed safeguarding, SAC should benefit from its off-policy nature, but its reliance on the probability of the chosen action outweighs this effect. This issue was most noticeable in the pendulum task, where the critic loss was continuously divergent. The poor initial performance in the quadrotor task could result from uninformative earlier samples, although the underlying reason for the drastic performance increase is unclear.
\par
PPO mostly benefitted from safeguarded training, especially in the balancing environments. There, safety is strongly tied to reward, enabling safeguarding to guide the exploration. In the energy system environment, boundary projection had a similar effect, however, ray masking significantly hindered learning. We attribute this to the diminished learning rate in the ray direction.
\par
The performance of SHAC with unaltered safeguards was mostly similar to unsafe training. Notable exceptions were the impaired learning in the quadrotor environment and the improved performance on the energy system task. The optimal action is mostly safe in balancing scenarios, such that the lack of gradient propagation in the mapping direction hurts learning. Due to the simplicity of the pendulum environment, the convergent runs showed barely any degradation compared to unsafe training. However, the increase in non-convergent runs on the pendulum with boundary projection is caused by a total loss of gradient information as outlined in \Cref{eq:boundary_projection_gradient}, since the action space in the pendulum environment is one-dimensional. In the more complex quadrotor task, the agents were learning very slowly or completely stalled for several individual runs. In contrast, the increase in return in the energy system environment could be due to the diminished gradients, which stabilise learning there.
\par
Our observations support \Cref{hyp:1} since the final policy and convergence speed of SHAC remained superior in the pendulum and quadrotor, while it narrowed the gap in the energy system.

\begin{table*}
  \centering
  \caption{Comparison of learning algorithms in safeguarded training.}
  \label{tab:safe}
  \begin{tabular}{l l r r r r r r}
    \toprule
    \multirow{2}{*}{Environment} & \multirow{2}{*}{Safeguard} & \multirow{2}{*}{Algorithm} & \multicolumn{2}{c}{\# Step} & \multicolumn{2}{c}{Return}           & \multirow{2}{*}{\# Stuck}                                                   \\
    \cmidrule(l){4-5}\cmidrule(l){6-7}
                                 &                            &                            & Mean                        & 95\% CI                              & Mean                      & 95\% CI                                &        \\
    \midrule
    \multirow{6}{*}{Pendulum}

                                 & \multirow{3}{*}{BP}        & SHAC                       & 23360                       & [\phantom{0}18560, \phantom{0}28800] & \textbf{-8}               & [\phantom{00000}-8, \phantom{00000}-8] & 2 / 10 \\
                                 &                            & SAC                        & 2504                        & [\phantom{00}2504, \phantom{00}2504] & -1083                     & [\phantom{00}-1103, \phantom{00}-1061] & 0 / 10 \\
                                 &                            & PPO                        & 80240                       & [\phantom{0}78880, \phantom{0}82960] & -10                       & [\phantom{0000}-10, \phantom{00000}-9] & 0 / 10 \\
    \cmidrule{2-8}
                                 & \multirow{3}{*}{RM}        & SHAC                       & 27392                       & [\phantom{0}20992, \phantom{0}33280] & \textbf{-8}               & [\phantom{00000}-8, \phantom{00000}-8] & 0 / 10 \\
                                 &                            & SAC                        & 2504                        & [\phantom{00}2504, \phantom{00}2504] & -424                      & [\phantom{000}-465, \phantom{000}-384] & 0 / 10 \\
                                 &                            & PPO                        & 76160                       & [\phantom{0}72080, \phantom{0}80240] & -12                       & [\phantom{0000}-12, \phantom{0000}-11] & 0 / 10 \\

    \midrule
    \multirow{6}{*}{Quadrotor}

                                 & \multirow{3}{*}{BP}        & SHAC                       & 45683                       & [\phantom{0}16498, \phantom{0}74854] & \textbf{-333}             & [\phantom{000}-394, \phantom{000}-265] & 0 / 10 \\
                                 &                            & SAC                        & 110176                      & [\phantom{0}90131, 123196]           & -338                      & [\phantom{000}-368, \phantom{000}-308] & 0 / 10 \\
                                 &                            & PPO                        & 118720                      & [\phantom{0}89600, 152320]           & -402                      & [\phantom{000}-453, \phantom{000}-350] & 0 / 10 \\
    \cmidrule{2-8}
                                 & \multirow{3}{*}{RM}        & SHAC                       & 67148                       & [\phantom{0}31909,\phantom{0} 99636] & \textbf{-251}             & [\phantom{000}-307, \phantom{000}-197] & 0 / 10 \\
                                 &                            & SAC                        & 80128                       & [\phantom{0}59081, 107171]           & -377                      & [\phantom{000}-415, \phantom{000}-337] & 0 / 10 \\
                                 &                            & PPO                        & 127680                      & [107520, 156800]                     & -379                      & [\phantom{000}-419, \phantom{000}-330] & 0 / 10 \\

    \midrule
    \multirow{6}{*}{Energy System}

                                 & \multirow{3}{*}{BP}        & SHAC                       & 366960                      & [213807, 509553]                     & -89167                    & [-111791, \phantom{0}-62775]           & 0 / 10 \\
                                 &                            & SAC                        & 491000                      & [290000, 685050]                     & -150179                   & [-252908, \phantom{0}-33560]           & 0 / 10 \\
                                 &                            & PPO                        & 661577                      & [457426, 905307]                     & \textbf{-2293}            & [\phantom{00}-3421, \phantom{000}-906] & 3 / 10 \\
    \cmidrule{2-8}
                                 & \multirow{3}{*}{RM}        & SHAC                       & 709280                      & [579920, 840433]                     & -8793                     & [\phantom{0}-12685, \phantom{00}-3679] & 0 / 10 \\
                                 &                            & SAC                        & 355999                      & [187974, 503073]                     & \textbf{-1843}            & [\phantom{00}-2396, \phantom{00}-1279] & 0 / 10 \\
                                 &                            & PPO                        & 548352                      & [313344, 801907]                     & -339006                   & [-485334, -197690]                     & 0 / 10 \\
    \bottomrule
  \end{tabular}
\end{table*}
\subsection{EVALUATION OF SAFEGUARDS}
\begin{figure*}[tp]
  \centering
  \includegraphics[width=\textwidth]{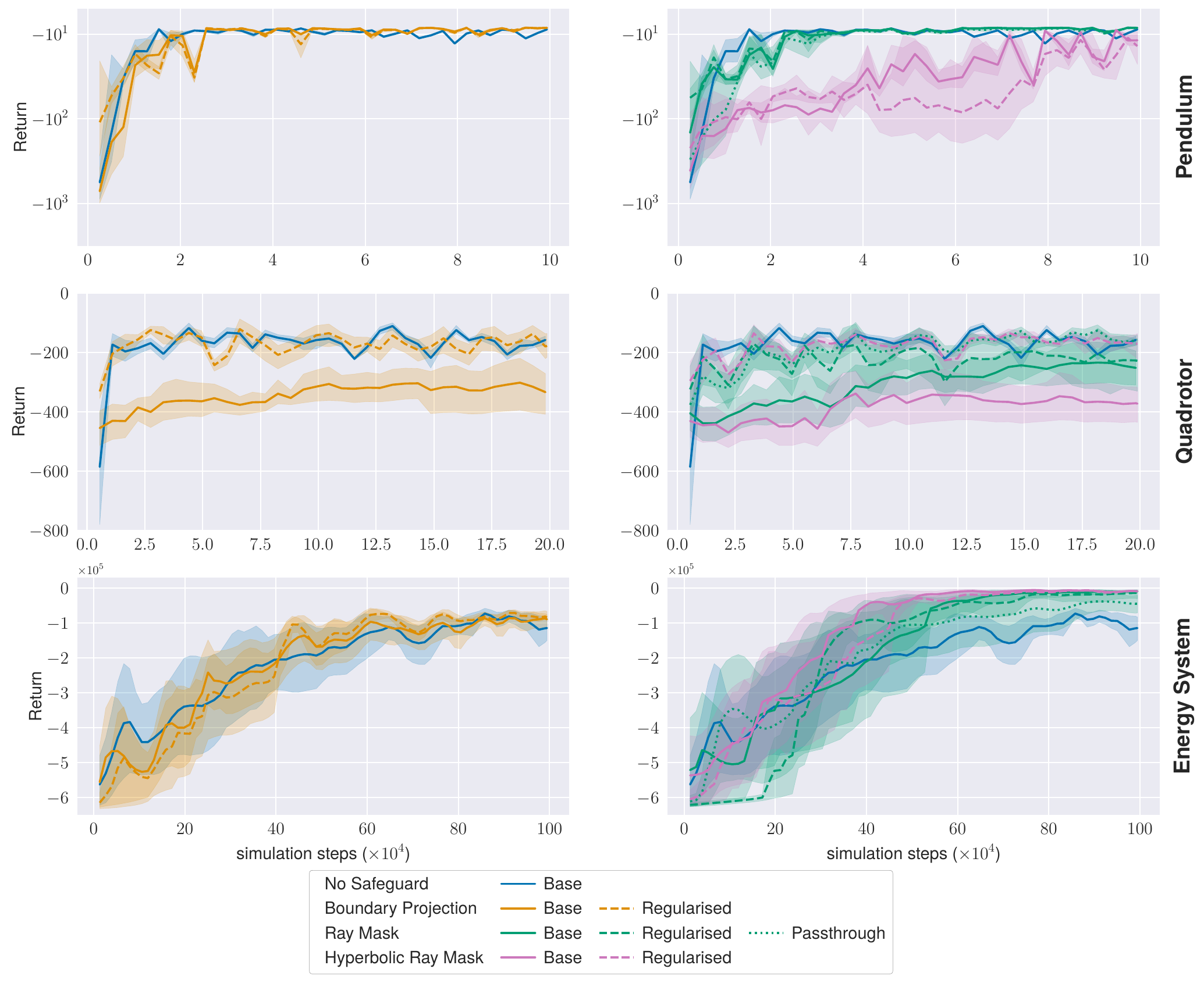}
  \caption{Comparison of SHAC under unsafe training and safeguarded training using boundary projection (left) and ray mask (right). Different line styles indicate the safeguard variants. In the pendulum environment, both base safeguards achieve comparable performance; in the quadrotor environment, regularised boundary projection and the regularised hyperbolic ray mask perform similarly; and in the energy system environment, all ray masks yield superior performance.}
  \label{fig:results}
\end{figure*}

Next, we evaluate \Cref{hyp:2} by comparing the safeguards introduced in \Cref{sec:method} to unsafe training on SHAC. \Cref{fig:results} shows the aggregated learning curves; we report the number of non-convergent runs in \Cref{tab:excluded}.
\par
For the unaltered safeguards, we observed a performance decline when the optimal action is safe compared to unsafe training. The impact was more severe for the unaltered boundary projection than for the ray mask, attributed to the lack of gradient propagation in the mapping direction. To this end, regularisation mostly improved the performance for both boundary projection and ray masking. Ray masking with a passthrough gradient and the hyperbolic ray mask had mixed results.
\begin{table}
  \centering
  \caption{Number of non-convergent runs for the various safeguards.}
  \label{tab:excluded}
  \begin{tabular}{l l r r r}
    \toprule
    \multicolumn{2}{c}{\multirow{2}{*}{Safeguard}} & \multicolumn{3}{c}{\# Stuck}                            \\
    \cmidrule(l){3-5}
                                                   &                              & Pen    & Quad   & ES     \\
    \midrule
    \multirow{2}{*}{BP}                            & Base                         & 2 / 10 & 0 / 10 & 0 / 10 \\

                                                   & Regularised                  & 1 / 10 & 0 / 10 & 0 / 10 \\
    \midrule
    \multirow{3}{*}{RM}                            & Base                         & 0 / 10 & 0 / 10 & 0 / 10 \\

                                                   & Regularised                  & 0 / 10 & 0 / 10 & 0 / 10 \\

                                                   & Passthrough                  & 2 / 10 & 1 / 10 & 2 / 10 \\
    \midrule
    \multirow{2}{*}{HRM}                           & Base                         & 1 / 10 & 0 / 10 & 0 / 10 \\

                                                   & Regularised                  & 0 / 10 & 0 / 10 & 0 / 10 \\
    \bottomrule
  \end{tabular}
\end{table}
\par
\textbf{Boundary projection with regularisation} alleviated most issues of the unaltered variant, as performance was on par with unsafe training. The observed reduction in non-convergent runs suggests that regularisation improves convergence. However, the fact that non-convergent runs persisted rather than being eliminated, indicates that the regularisation coefficient may be too small. The observation that non-convergent runs involve more safeguarding interventions than convergent ones supports this assumption. Due to time constraints, we could not run additional experiments with an increased regularisation coefficient.
\par
\textbf{Ray masking with regularisation} had less changes, as it improved the convergence speed in the quadrotor environment, but not to the level of unsafe training. We attribute the negligible effect to the fact that regularising the ray mask constantly introduces a gradient towards the centre. In contrast, regularisation only influences policy updates when actions are unsafe for boundary projection.
\par
\textbf{Ray masking with a passthrough gradient} can improve performance since a robust control invariant state set captures most of the optimal actions in balancing tasks, which retains the gradient correctness while eliminating the gradient decrease in the mapping direction. Since the unaltered ray mask is almost optimal in the pendulum environment, performance increases are only visible in the quadrotor environment, where the gap to unsafe training is closed. The non-convergence of some runs in the balancing tasks could be due to the effectively increased learning rate, as the gradient is no longer diminished by the safeguard. In the energy system task, not all reward-maximising actions are safe, such that the gradients of this safeguard can be wrong and therefore stall learning.
\par
The \textbf{hyperbolic ray mask} produces a similar mapping distance to boundary projection due to the hyperbolic tangent function, leading us to expect comparable performance. Unlike boundary projection, the hyperbolic map ensures that a gradient is always available. However, for unsafe actions, this gradient remains small. We observe marginally more stable convergence but significantly lower policy quality than boundary projection in the balancing tasks. This result is unexpected and may be attributed to the diminished gradient in the mapping direction, as indicated by the frequent safeguarding interventions. The performance of the regularised, hyperbolic ray mask supports this statement, as it achieved the best performance in the quadrotor environment and converged in all ten runs. The large confidence interval and poor mean performance in the pendulum environment were attributed to a single outlier, which converged significantly slower than all other runs.

\subsection{COMPARISON OF SAFE CENTRE APPROXIMATIONS}
We also compare safe centre approximations for the ray mask, where we found that the zonotopic approximation results in superior final policies and faster convergence, see \Cref{tab:orp}. Since the orthogonal approximation only applied to unsafe actions, safe actions were not mapped. In the pendulum task, the one-dimensional action space allows for exact safe centre approximations, which condenses the comparison to rarer interventions by the orthogonal approximation versus the smoother map of the zonotopic approximation. The continuous map provided by the zonotopic approximation produced superior final policies and converged faster. The low number of steps of the orthogonal approximation in the quadrotor task was an artefact of the worse policy, as seen in the learning curves in \Cref{app:orp_vs_zrp}. In the energy system, the orthogonal approach converged faster initially, but to a worse policy. Therefore, smooth safeguards appear more critical than rare interventions for learning.
\begin{table*}
  \centering
  \caption{Comparison of the safe centre approximations.}
  \label{tab:orp}
  \begin{tabular}{l l r r r r r}
    \toprule
    \multirow{2}{*}{Environment} & \multirow{2}{*}{Approximation} & \multicolumn{2}{c}{\# Step} & \multicolumn{2}{c}{Return}           & \multirow{2}{*}{\# Stuck}                                                 \\
    \cmidrule(l){3-4}\cmidrule(l){5-6}
                                 &                                & Mean                        & 95\% CI                              & Mean                      & 95\% CI                              &        \\

    \midrule
    \multirow{2}{*}{Pendulum}

                                 & Zonotopic                      & 27392                       & [\phantom{0}21241, \phantom{0}32768] & -8                        & [\phantom{0000}-8, \phantom{0000}-8] & 0 / 10 \\
                                 & Orthogonal                     & 30208                       & [\phantom{0}18432, \phantom{0}39424] & -8                        & [\phantom{0000}-8, \phantom{0000}-8] & 0 / 10 \\

    \midrule
    \multirow{2}{*}{Quadrotor}

                                 & Zonotopic                      & 67148                       & [\phantom{0}30808, 101287]           & \textbf{-251}             & [\phantom{00}-306, \phantom{00}-194] & 0 / 10 \\
                                 & Orthogonal                     & 31372                       & [\phantom{00}5476, \phantom{0}57241] & -432                      & [\phantom{00}-482, \phantom{00}-371] & 0 / 10 \\

    \midrule
    \multirow{2}{*}{Energy System}

                                 & Zonotopic                      & 709280                      & [579898, 829873]                     & \textbf{-8793}            & [-12852, \phantom{0}-3797]           & 0 / 10 \\
                                 & Orthogonal                     & 109560                      & [\phantom{0}79200, 138632]           & -79594                    & [-88277, -70587]                     & 0 / 10 \\
    \bottomrule
  \end{tabular}
\end{table*}

\subsection{COMPARISON OF COMPUTATION TIME}
The relative computation time of safeguarded training is compared to its unsafe counterpart in \Cref{tab:wallclock} to estimate computational overhead. For this purpose, the computation time was measured over 10,000 steps in the pendulum environment. The results show at least a four-fold increase in computation time when boundary projection is applied. Ray masking took almost double the time of boundary projection, which we trace to the increased computational complexity of its optimisation problems, due to the derived safe action set. SHAC produced around a quarter of the additional computational overhead, as it must maintain the computational graph for backpropagation. The increased computation time poses a significant downside, although a custom, more efficient implementation could mitigate the effects. Moreover, safeguarding via a ray mask is significantly cheaper for specified safe action sets, since the safe centre is provided. However, pre-computing the safe state or action set may not be possible depending on the task, which could further increase the computation needed per training iteration.
\begin{table}
  \centering
  \caption{Relative computation time of the different safeguards for 10,000 steps in the pendulum environment compared to their unsafe versions.}
  \label{tab:wallclock}
  \begin{tabular}{l>{\centering\arraybackslash}m{1.3cm} >{\centering\arraybackslash}m{1.3cm} >{\centering\arraybackslash}m{1.3cm}}
    \toprule
                        & \multicolumn{3}{c}{Learning Algorithm}                 \\
    \cmidrule(l){2-4}
    Safeguard           & SHAC                                   & PPO   & SAC   \\
    \midrule
    No Safeguard        & 1.000                                  & 1.000 & 1.000 \\
    Boundary Projection & 5.089                                  & 4.483 & 4.774 \\
    Ray Mask            & 9.857                                  & 8.100 & 7.500 \\
    \bottomrule
  \end{tabular}
\end{table}
\section{LIMITATIONS AND CONCLUSION}
This work demonstrates the fundamental applicability and effectiveness of safeguards for analytic gradient-based reinforcement learning, unlocking its usage for safely training agents in simulations before deploying them in safety-critical applications. While we showcased the possibility of achieving performance on par with or exceeding unsafe training, success depends on the quality and representation of the safe set.
\par
While we utilised zonotopes, the safeguards presented are not limited to this set representation. The only limitation of the representation is the star-shapedness of the ray mask and the ability to solve the relevant containment problems in an efficient and differentiable manner. The general trade-off in the choice of set representation is achieving a tight approximation of the true safe set versus computationally cheap containment problems. A limitation of the chosen zonotope representation is its inherent symmetry, which can unnecessarily restrict the safe set. This restriction becomes apparent near the boundary of $\mathcal{A}$, where the minimum distance to the safe set boundary constrains the extent of the set both towards and away from it.
\par
Moreover, using CVXPY allows for rapid prototyping but may not offer optimal performance compared to custom solvers and formulations, which could decrease the substantial overhead of safeguarding. The group behind CVXPY recently addressed this issue by a parallel interior point solver \cite{cuClarabel} and CVXPYgen \cite{cvxpygen}, which generates a custom solver in C.
\par
In general, safeguarding for the sole sake of efficiency requires either an informative, safe set or an expensive simulation since the sample efficiency gains strongly depend on the quality of the safe set. In contrast, the computational overhead depends only on the representation and dimensionality of the safe action set.
\par
The presented safeguards worked well but are likely not optimal. An interesting idea for future work is deriving a general bijective map inspired by the ray mask \cite{RayMasking} and gauge map \cite{Gauge}. This map would again project actions radially towards an interior point of the safe action set, but the optimal interior point could be different from the geometric centre. Different projection centres could be advantageous in cases where the safe action set is adjacent to the corner of the feasible set, which would shrink the space unevenly. In addition, optimising the trade-off between the mapping distance and the gradient strength could improve convergence properties.

\printbibliography

\begin{IEEEbiography}[{\includegraphics[width=1in,height=1.25in, clip,keepaspectratio]{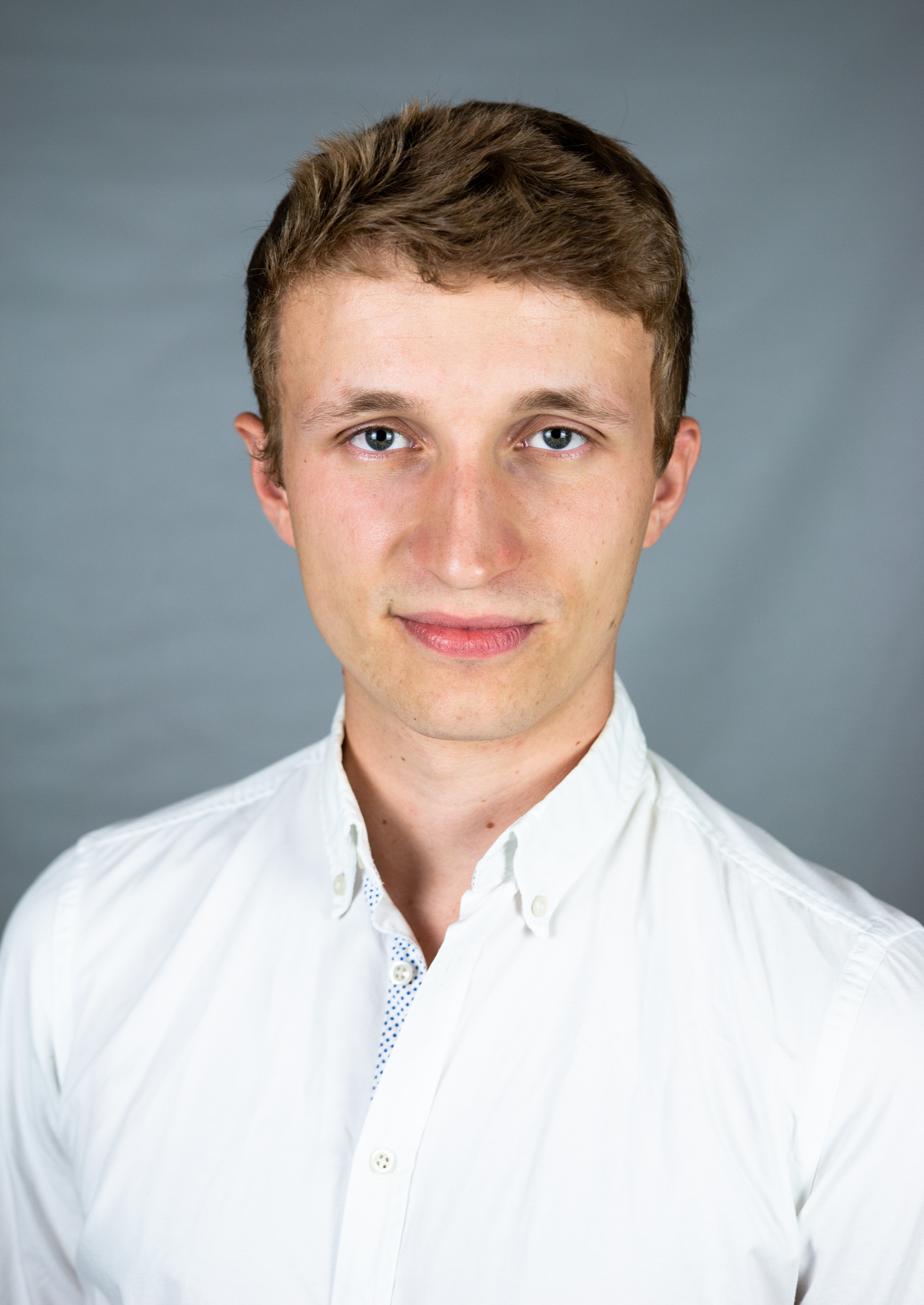}}]{T. Walter}{\space} (Member, IEEE) received the B.Eng. degree in Electrical Engineering and Information Technology from the University of Applied Sciences Munich, Munich, Germany, in 2022, and the M.Sc. degree in Computational Science and Engineering (Honour's track) from the Technical University of Munich, Munich, Germany, in 2025.
  He joined the Cyber-Physical Systems Group at the Technical University of Munich in 2025. His research interests include modular robotics, machine learning, optimisation, and control theory with manufacturing applications.
\end{IEEEbiography}
\begin{IEEEbiography}[{\includegraphics[width=1in,height=1.25in,clip,keepaspectratio]{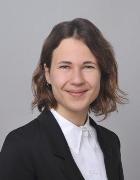}}]
  {Hannah Markgraf}{\space}received a B.Sc. degree in mechanical engineering in 2019 and an M.Sc. in automation and control in 2021, both from RWTH Aachen University. She is working toward a PhD in computer science at the Cyber-Physical Systems Group at the Technical University of Munich. Her research interests include reinforcement learning and optimisation with application to energy management systems.
\end{IEEEbiography}
\begin{IEEEbiography}[{\includegraphics[width=1in,height=1.25in,clip,keepaspectratio]{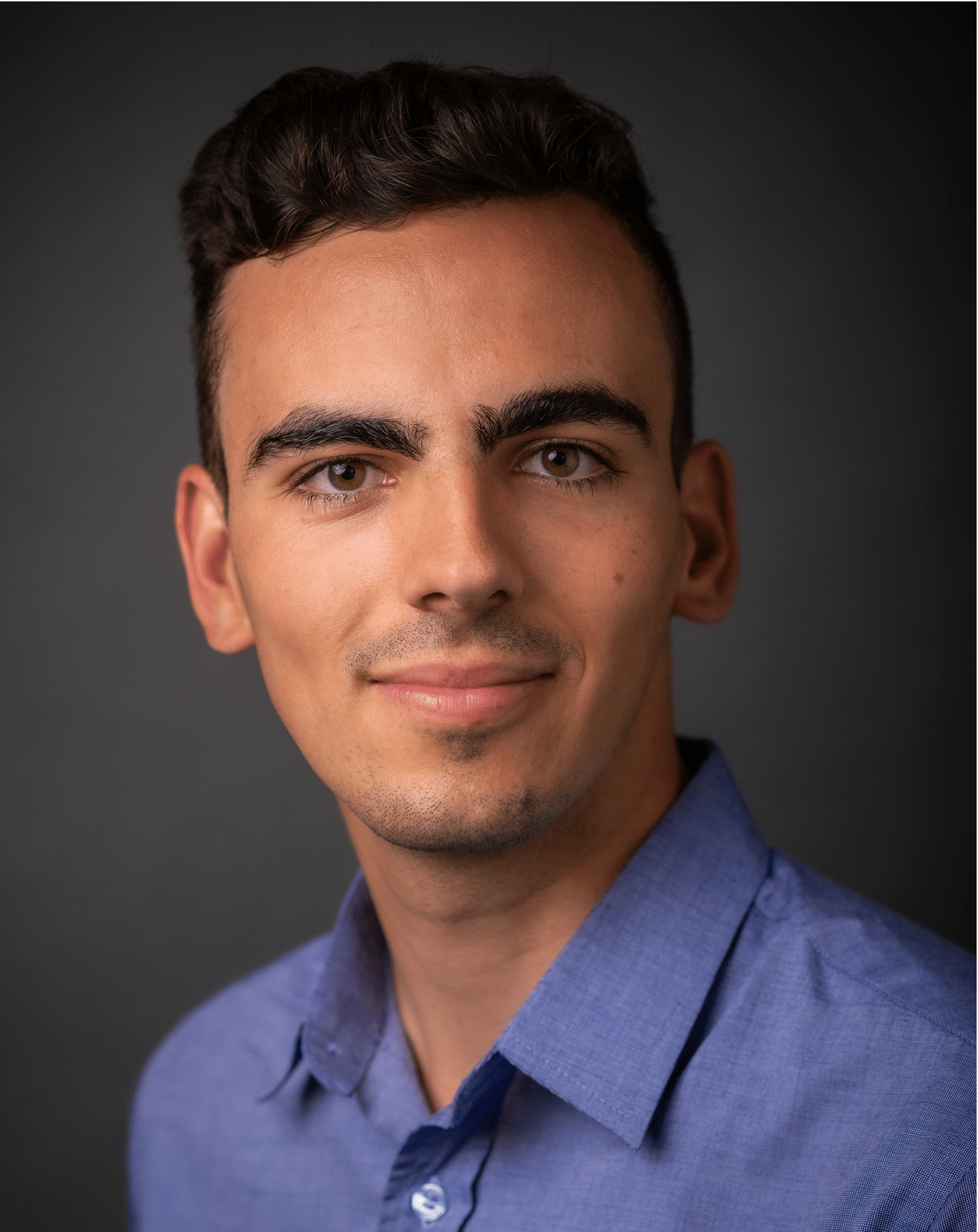}}]
  {Jonathan K\"ulz}{\space}received a B.Sc. degree in mechatronics and information technology in 2017 from Karlsruhe Institute of Technology and an M.Sc. degree in robotics, cognition and intelligence in 2021 from Technische Universität München, Munich, Germany, where he is currently working toward the Ph.D. degree in computer science.
  His research interests include robot morphology optimisation and computational co-design.
\end{IEEEbiography}
\begin{IEEEbiography}[{\includegraphics[width=1in,height=1.25in,clip,keepaspectratio]{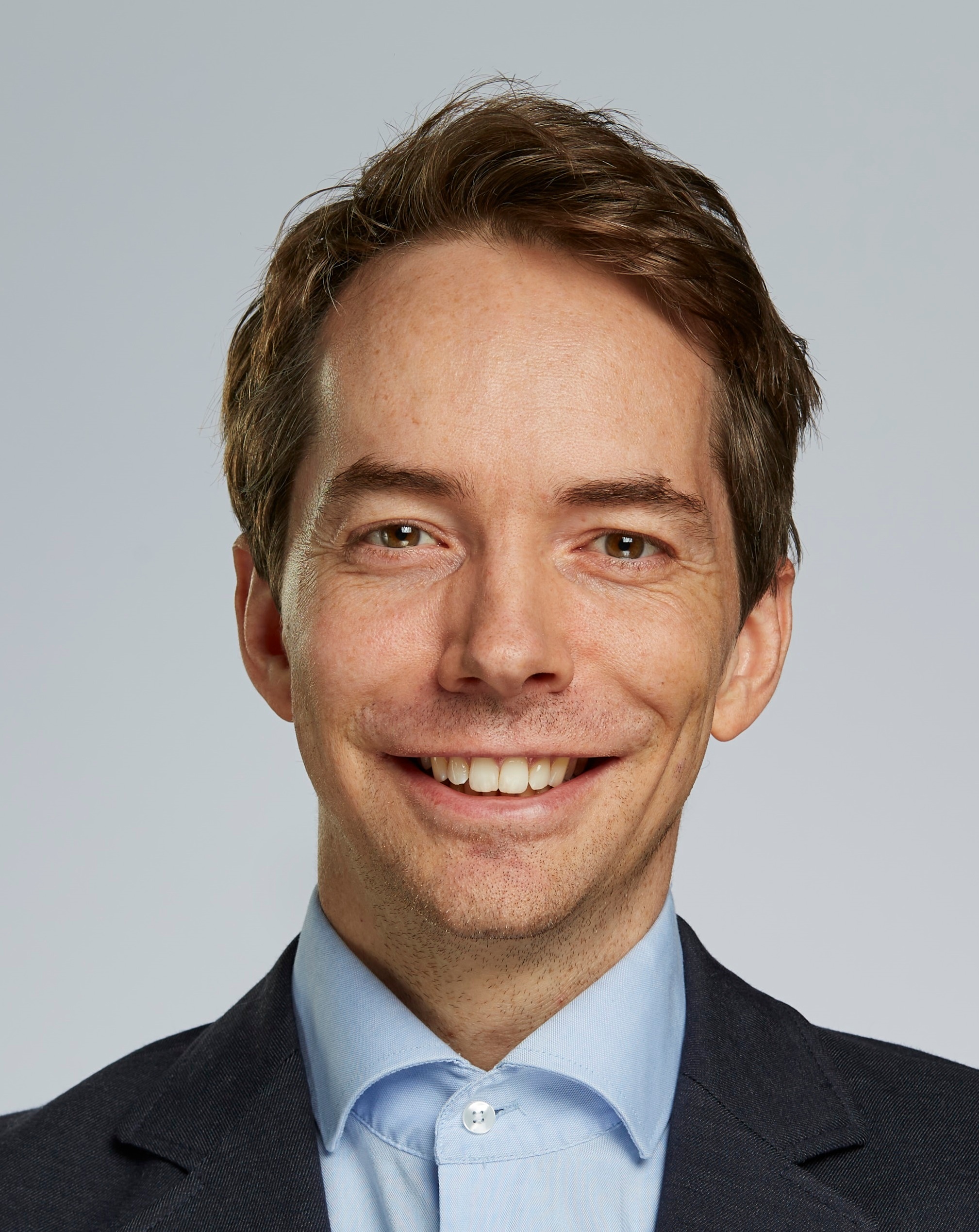}}]
  {Matthias Althoff}{\space} is an associate professor in computer science at the Technical University of Munich, Germany. He received his diploma engineering degree in Mechanical
  Engineering in 2005, and his PhD degree in Electrical Engineering in
  2010, both from the Technical University of Munich, Germany.
  From 2010 to 2012, he was a postdoctoral researcher at Carnegie Mellon University, Pittsburgh, USA, and from 2012 to 2013, an assistant professor at Technische Universit\"at Ilmenau, Germany. His research interests include formal verification of continuous and hybrid systems, reachability analysis, planning algorithms, nonlinear control, automated vehicles, and power systems.
\end{IEEEbiography}

\section*{APPENDIX}
\addcontentsline{toc}{section}{APPENDIX} 

\setcounter{subsection}{0}%
\renewcommand\thesubsection{\Alph{subsection}}
\renewcommand\theHsubsection{\Alph{subsection}}%

\crefname{subsection}{Appendix}{Appendices}%
\Crefname{subsection}{Appendix}{Appendices}%
\crefalias{subsection}{appendix}%

\subsection{ENVIRONMENT DESCRIPTIONS} \label{app:envs}
All environments share some characteristics: the feasible state and action sets are axis-aligned boxes; the feasible action set is of unit length; and the dynamics is given as a first-order ordinary differential equation, which we integrate using an Euler scheme. The non-determinism of the system is encapsulated by additive bounded noise. This yields the transition function
\begin{equation}
  s_{i+1} = s_i +dt \left(\dot{s}_i + w_i\right)
\end{equation}
with the discrete time step size $dt$ and the noise sample $w_i \in \mathcal{W}$, where $\mathcal{W}$ is a zonotope. For the energy system, we utilise an explicit Euler scheme, whereas for the pendulum and quadrotor, we use a semi-implicit Euler scheme
\begin{equation}
  s_{i+1} = \begin{bmatrix}
    p_i \\ \dot{p}_{i}
  \end{bmatrix} +dt \left(\begin{bmatrix}
      \dot{p}_{i+1} \\ \ddot{p}_{i}
    \end{bmatrix} + w_i\right) \, ,
\end{equation}
where we exploit the form of our state $s=\begin{bmatrix}
    p \\ \dot{p}
  \end{bmatrix}$, which consists only of coordinates $p$ and their respective velocities $\dot{p}$. We choose a semi-implicit Euler integrator, since it is symplectic.
\par
\textbf{Pendulum} This environment possesses a feasible state set $S =\left[-\pi, \pi\right] \times\left[-8, 8\right]$ with the state $s = \begin{bmatrix} \theta & \dot{\theta} \end{bmatrix}^T$ representing the angle $\theta$ and the angular velocity $\dot{\theta}$. The feasible action set has one dimension with the action $a$ representing the torque. The dynamics is
\begin{equation}
  \dot{s}(s, a) = \begin{bmatrix}
    \dot{\theta}                                    \\
    \frac{1.5g\sin{\theta}}{l}+\frac{3ca}{ml^2} + w \\
  \end{bmatrix}
\end{equation}
with the gravitational acceleration $g$, the length $l$, the mass $m$, and the torque magnitude $c$. The noise zonotope is $\mathcal{W} = \left<\begin{bmatrix}
    0 \\ 0
  \end{bmatrix}, \begin{bmatrix}
    0   & 0 \\
    0.1 & 0
  \end{bmatrix}\right>$. The reward function is
\begin{equation}
  r(s, a) = -\theta^2 - \frac{\dot{\theta}^2}{10} - \frac{a^2}{100}
\end{equation}
and encodes the goal of balancing the pendulum upright. Colloquially, we define safety as the part of the state space from which the controller can maintain balance without the pendulum falling. Formally, we derive a safe action set from a robust control invariant (RCI) state set, which we obtain by the method in Schäfer et al. \cite{RCI-Sets}.
\par
\textbf{Quadrotor} This environment possesses a feasible state set
\begin{align}
  \mathcal{S} = [-8, 8]^2 & \times \left[-\frac{\pi}{12}, \frac{\pi}{12}\right] \times [-0.8,0.8]       \\
                          & \times [-1.0, 1.0] \times \left[-\frac{\pi}{2}, \frac{\pi}{2}\right] \notag
\end{align}
with the state $s = \begin{bmatrix} x & y & r & \dot{x} & \dot{y} & \dot{r} \end{bmatrix}^T$ representing the quadrotor position $(x, y)$, roll $r$, and their respective velocities $(\dot{x}, \dot{y}, \dot{r})$. The feasible action set has two dimensions with the thrust $a_0$ and roll angle $a_1$. The dynamics is
\begin{equation}
  \dot{s} = \begin{bmatrix}
    \dot{x}                           \\
    \dot{y}                           \\
    \dot{r}                           \\
    (a_0 c_{0} + g) \sin{r} + w_0     \\
    (a_0 c_{0} + g) \cos{r} - g + w_1 \\
    a_1 c_{1} pd_2 - pd_0 r - pd_1 \dot{r}\end{bmatrix}
\end{equation}
with the torque magnitude $c_0$, the roll angle magnitude $c_1$, and the PID gains $pd_{0-2}$. The noise zonotope is $\mathcal{W} = \left<\textbf{0}, \begin{bmatrix}
    0.1 & 0.1 & 0 & 0 & 0 & 0
  \end{bmatrix}_D\right>$. The reward is
\begin{equation}
  \begin{split}
    r(s, a) & = -2.5 \sqrt{(x - x_0)^2 + (y - y_0)^2}                          \\
            & - \frac{r+\dot{x}+\dot{y}+\dot{r}}{10} - \frac{(a_0c_0+g)^2}{50} \\
            & - \frac{(a_1c_1)^2}{100} \, ,
  \end{split}
\end{equation}
where $x_0, y_0$ encode the initial position of the quadrotor. The reward function encodes balancing the quadrotor around its initial location. Again, we derive a safe action set from an RCI set to obtain the set of safe actions.
\par

\textbf{Energy Management System}
This system has the feasible state set $\mathcal{S} = [0, 10] \times [18, 24] \times [10, 100]$ with the state
$s= \begin{bmatrix} e & \vartheta^\text{in} & \vartheta^\text{ret}\end{bmatrix}^T$, where $e$ is the charge of the battery, $\vartheta_\text{in} $ is the indoor temperature of the building, and $\vartheta_\text{ret} $ is the return temperature of the floor heating system. The feasible action set has two dimensions, representing the power set point for the battery $a_0$ and the heat pump $a_1$. The dynamics is
\begin{equation}
  \dot{s} = \begin{bmatrix}
    a_0                                                 \\
    -c_0 \vartheta^\text{in} + c_1 \vartheta^\text{ret} \\
    c_2 \vartheta^\text{in} - c_2 \vartheta^\text{ret} + c_3 a_1
  \end{bmatrix},
\end{equation}
where the computation of the coefficients $c_{1-3}$ is detailed in \cite[Eq. 2.17]{bianchi2005dissertation}. The noise zonotope is
\begin{equation}
  \mathcal{W} = \left<\begin{bmatrix}
    0 \\ 10.8986\\ 0
  \end{bmatrix}, \begin{bmatrix}
    0       & 0 & 0 \\
    21.1985 & 0 & 0 \\
    0       & 0 & 0
  \end{bmatrix}\right> \,
\end{equation}
resulting from the replayed data for the outdoor temperature. The reward is
\begin{align}
  r(s,a) = & - (a_0 + a_1 + p^\ell - p^\text{PV}) dt \, \phi                  \\
           & - 100 (\vartheta^\text{in} - \vartheta^\text{set})^2 \notag \, ,
\end{align}
where $p^\ell$ is the inflexible load of the building, $p^\text{PV}$ is the output of the photovoltaic generator, $\phi$ is the electricity price, and $\vartheta^\text{set}$ is the desired indoor temperature. The reward function encodes the goal of minimising energy consumption while maintaining room temperature. To facilitate the task, the observation $o_t = \begin{bmatrix}
    s_t & \vartheta^\text{out}_{[t:t+H]} & p^\text{PV}_{[t:t+H]} & p^\ell_{[t:t+H]} & \phi_{[t:t+H]}
  \end{bmatrix}^T$ includes the outdoor temperature $\vartheta^\text{out}$, current measurements, and forecasts, where we use the slicing notation $x_{\left[i:j\right]} = \begin{bmatrix}
    x_i & \dots & x_{j-1}
  \end{bmatrix}^T$.
We choose $H=5$, resulting in $23$ observations. The safe state set is the feasible state set.

\subsection{PROOF OF THEOREM 1: JACOBIAN OF BOUNDARY PROJECTION} \label{app:jacobian}
\begin{IEEEproof}
  We investigate the rank of the Jacobian of boundary projection \Cref{eq:boundary_projection}, namely $\text{rank}(\frac{\partial g_{\text{BP}}(a)}{\partial a}) = \text{rank}(\frac{\partial a_s}{\partial a})$, by utilising the differentials of the KKT conditions of a canonical, quadratic program \cite[Eq. 6]{DifferentiateQuadraticProgram}
  \begin{equation}
    \label{eq:differentials}
    \begin{aligned}
      \begin{bmatrix}
        Q            & K^T         & A^T        \\
        \lambda_D^*K & (Kz^*- h)_D & \textbf{0} \\
        A            & \textbf{0}  & \textbf{0}
      \end{bmatrix}
      \begin{bmatrix} \diff z\\\diff \lambda\\ \diff \nu \end{bmatrix} = \\
      \begin{bmatrix}
        -\diff Qz^* - \diff q - \diff K^T\lambda^* - \diff A^T\nu^* \\
        -\lambda_D^*\diff Kz^* + \lambda_D^*\diff h                 \\
        -\diff Az^* + \diff b
      \end{bmatrix}\,
    \end{aligned}
  \end{equation}
  where the superscript $*$ denotes optimal values, bold scalars a constant matrix of suitable size with all entries equal to the scalar, $\nu \in \mathbb{R}^d$ the dual variables on the equality constraints, $\lambda \in \mathbb{R}^{2n}$ the dual variables on the inequality constraints, and $\diff$ a differential. Under the assumptions in \Cref{lem:boundary_projection}, \Cref{eq:boundary_projection} is
  \begin{subequations}
    \label{eq:quad_bp}
    \begin{alignat}{2}
       & \min_{a_{s}, \gamma} & \norm{a-a_s}^2                                    \\
       & \text{subject to }   & a_s = c_{\mathcal{A}_s} + G_{\mathcal{A}_s}\gamma \\
       &                      & \anynorm{\gamma}_\infty \leq 1
    \end{alignat}
  \end{subequations}
  which we reformulate in canonical, quadratic form
  \begin{subequations}
    \begin{alignat}{2}
       & \min_{z}          & \frac{1}{2}z^TQz+q^Tz                   \\
       & \text{subject to} & Az=b                                    \\
       &                   & Kz \leq h \label{constraint:inequality}
    \end{alignat}
  \end{subequations}
  with
  \begin{align}
    z & = \begin{bmatrix} a_s \\ \gamma \end{bmatrix} \in \mathbb{R}^{d+n} \label{eq:z} \\
    Q & = \begin{bmatrix}
            2 I_d      & \textbf{0} \\
            \textbf{0} & \textbf{0} \\
          \end{bmatrix} \in \mathbb{R}^{(d+n) \times (d+n)} \label{eq:Q}                \\
    q & = -2 \begin{bmatrix}
               a \\ \textbf{0}
             \end{bmatrix}\in \mathbb{R}^{d+n} \label{eq:q}                             \\
    A & = \begin{bmatrix}
            I_d & -G_{\mathcal{A}_s}
          \end{bmatrix}\in \mathbb{R}^{d \times (d+n)} \label{eq:A}                     \\
    b & = c_{\mathcal{A}_s} \in \mathbb{R}^{d}         \label{eq:b}                     \\
    K & = \begin{bmatrix}
            \textbf{0} & I_n  \\
            \textbf{0} & -I_n
          \end{bmatrix} \in \mathbb{R}^{2n \times (d+n)}    \label{eq:K}                \\
    h & = \textbf{1} \in \mathbb{R}^{2n}       \label{eq:h}  \, ,
  \end{align}
  where the subscript of the identity denotes its size. We remark the equality of the objectives $\min_{a_{s}}  \norm{a-a_s}^2 = \min_{a_{s}} a_s^Ta_s - 2a^Ta_s$, since the remaining term $a^Ta$ is independent of $a_s$ and we are only interested in the minimiser $a_s$. To obtain the Jacobian with respect to the action, we substitute $\diff q \overset{(\ref{eq:q})}{=} -2 \begin{bmatrix} I_d\\ \textbf{0} \end{bmatrix}$ and all other differential terms with zero, as they are independent of $a$, leaving \Cref{eq:differentials} as
  \begin{equation}
    \begin{bmatrix}
      Q            & K^T        & A^T        \\
      \lambda_D^*K & (Kz^*-h)_D & \textbf{0} \\
      A            & \textbf{0} & \textbf{0}
    \end{bmatrix}
    \begin{bmatrix}  \frac{\partial z}{\partial a} \\\frac{\partial \lambda}{\partial a} \\\frac{\partial \nu}{\partial a} \end{bmatrix} =
    \begin{bmatrix}
      2 \begin{bmatrix} I_d\\\textbf{0} \end{bmatrix} \\
      \textbf{0}                                      \\
      \textbf{0}
    \end{bmatrix} \, .
  \end{equation}
  We insert \Cref{eq:z,eq:Q,eq:A,eq:K,eq:h}, such that $z^* = \begin{bmatrix}
      a_s^* & \gamma^*
    \end{bmatrix}^T$ and $\frac{\partial z}{\partial a} = \begin{bmatrix}
      \frac{\partial a_s}{\partial a} & \frac{\partial \gamma}{\partial a}
    \end{bmatrix}^T$, which expands the system into
  \begin{align} \label{sys:main}
     & \begin{bmatrix}
         2I_d       & \textbf{0}                                         & \textbf{0}                                 & I_d                  \\
         \textbf{0} & \textbf{0}                                         & {\begin{bmatrix}I_n & -I_n  \end{bmatrix}} & -G_{\mathcal{A}_s}^T \\
         \textbf{0} & \lambda_D^*\begin{bmatrix}I_n \\ -I_n\end{bmatrix} & \begin{bmatrix}
                                                                          \gamma_D^* -I_n & \textbf{0}       \\
                                                                          \textbf{0}      & -\gamma_D^* -I_n
                                                                        \end{bmatrix}      & \textbf{0}                                \\
         I_d        & -G_{\mathcal{A}_s}                                 & \textbf{0}                                 & \textbf{0}
       \end{bmatrix}
    \begin{bmatrix}
      \frac{\partial a_s}{\partial a}     \\
      \frac{\partial \gamma}{\partial a}  \\
      \frac{\partial \lambda}{\partial a} \\
      \frac{\partial \nu}{\partial a}     \\
    \end{bmatrix} \nonumber                                                             \\
     & \qquad =
    \begin{bmatrix}
      2I_d       \\
      \textbf{0} \\
      \textbf{0} \\
      \textbf{0} \\
    \end{bmatrix}\,.
  \end{align}
  This yields a coupled system of matrix equations
  \begin{align}
    2\frac{\partial a_s}{\partial a}+\frac{\partial \nu}{\partial a}                                                                 & = 2I_d \label{sys:1}                                        \\
    {\begin{bmatrix}I_n & -I_n  \end{bmatrix}}\frac{\partial \lambda}{\partial a}-G_{\mathcal{A}_s}^T\frac{\partial \nu}{\partial a} & = \textbf{0} \label{sys:2}                                  \\
    \lambda_D^*\begin{bmatrix}I_n \\ -I_n\end{bmatrix}\frac{\partial \gamma}{\partial a}+\begin{bmatrix}
                                                                                           \gamma_D^* -I_n & \textbf{0}       \\
                                                                                           \textbf{0}      & -\gamma_D^* -I_n
                                                                                         \end{bmatrix}\frac{\partial \lambda}{\partial a}                             & = \textbf{0} \label{sys:3} \\
    \frac{\partial a_s}{\partial a}-G_{\mathcal{A}_s}\frac{\partial \gamma}{\partial a}                                              & = \textbf{0} \label{sys:4}\, .
  \end{align}
  We solve this system by substitution, starting from \Cref{sys:4}:
  \begin{equation} \label{sys:4_reordered}
    \frac{\partial a_s}{\partial a} = G_{\mathcal{A}_s}\frac{\partial \gamma}{\partial a} \, ,
  \end{equation}
  which we substitute into \Cref{sys:1}
  \begin{equation}
    \frac{\partial \nu}{\partial a} = 2I_d - 2G_{\mathcal{A}_s}\frac{\partial \gamma}{\partial a} \, .
  \end{equation}
  We substitute this into \Cref{sys:2}
  \begin{equation}\label{sys:relate_lambda_gamma}
    \frac{1}{2}{\begin{bmatrix}I_n & -I_n  \end{bmatrix}}\frac{\partial \lambda}{\partial a} = G_{\mathcal{A}_s}^T - G_{\mathcal{A}_s}^TG_{\mathcal{A}_s}\frac{\partial \gamma}{\partial a} \, .
  \end{equation}
  To utilise the remaining \Cref{sys:3}, we partition it by the activity of the inequality constraint in \Cref{constraint:inequality} utilising the KKT complementarity slackness conditions \cite[Eq. 4.3]{DifferentiateQuadraticProgram} in element-wise notation
  \begin{equation}
    \forall i = 1, \ldots, n:\lambda_i^*(K_{i,:} z_* - h_i) = 0  \, .
  \end{equation}
  The constraint is inactive $\lambda_i^* = 0$ or active $\gamma_i^*=\pm1$, where we write $\pm$ for shortness to denote the lower and upper part of the supremum norm. We define the index set of active constraints as $\mathcal{I}_a = \left\{i \mid \gamma_i^*=\pm\textbf{1}\right\}$ and of inactive constraints as $\mathcal{I}_i = \left\{i \mid \lambda_i^*=0\right\}$. We reorder the generators according to the index sets and partition \Cref{sys:3} into
  \begin{equation}\label{sys:3_part}
    \begin{bmatrix}\left(\pm\lambda^*_{\mathcal{I}_a}\right)_D\left(\frac{\partial \gamma}{\partial a}\right)_{\mathcal{I}_a, :} \\
      (\pm\gamma^*_{\mathcal{I}_i} - \textbf{1})_D\left(\frac{\partial \lambda}{\partial a}\right)_{\mathcal{I}_i, :}
    \end{bmatrix} = \begin{bmatrix}
      \textbf{0} \\ \textbf{0}
    \end{bmatrix} \, ,
  \end{equation}
  where we utilise the diagonalised form of the optimal variables and denote full rows or columns with a colon subscript. Using the assumption of strict complementarity we have
  \begin{align}
     & \forall i \in \mathcal{I}_a: \lambda^*_i > 0             \\
     & \forall i \in \mathcal{I}_i:\pm\gamma^*_i - 1 \neq 0\, ,
  \end{align}
  which means the diagonal matrices $\left(\lambda^*_{\mathcal{I}_a}\right)_D$ and $(\pm\gamma^*_{\mathcal{I}_i} - \textbf{1})_D$ are both invertible, yielding
  \begin{align}
    \left(\frac{\partial \gamma}{\partial a}\right)_{\mathcal{I}_a,:}   & = \textbf{0} \label{res:gamma}      \\
    \left(\frac{\partial \lambda}{\partial a}\right)_{\mathcal{I}_i, :} & = \textbf{0}\, . \label{res:lambda}
  \end{align}
  To obtain the remaining part $\left(\frac{\partial \gamma}{\partial a}\right)_{\mathcal{I}_i,:}$, we examine the inactive rows in \Cref{sys:relate_lambda_gamma} and substitute \Cref{res:lambda} yielding
  \begin{equation}
    \left(G^T_{\mathcal{A}_s}\right)_{\mathcal{I}_i, :}\left(G_{\mathcal{A}_s}\right)_{:, \mathcal{I}_i}\left(\frac{\partial \gamma}{\partial a}\right)_{\mathcal{I}_i,:} = \left(G_{\mathcal{A}_s}^T\right)_{\mathcal{I}_i,:} \, ,
  \end{equation}
  which is always solvable, since the right-hand side lies trivially in the column space of the left-hand side. We utilise the Moore-Penrose inverse as a solution
  \begin{equation}
    \label{eq:inactive_solution}
    \left(\frac{\partial \gamma}{\partial a}\right)_{\mathcal{I}_{i,:}}   =\left(\left(G_{\mathcal{A}_s}^T\right)_{\mathcal{I}_i,:}\left(G_{\mathcal{A}_s}\right)_{:,\mathcal{I}_i}\right)^{\dagger}\left(G_{\mathcal{A}_s}^T\right)_{\mathcal{I}_i,:}
  \end{equation}
  and combine \Cref{res:gamma} and \Cref{eq:inactive_solution} into
  \begin{equation}
    \label{eq:combined_res}
    \frac{\partial \gamma}{\partial a} = \begin{bmatrix}
      \textbf{0} \\ \left(\left(G_{\mathcal{A}_s}^T\right)_{\mathcal{I}_i,:}\left(G_{\mathcal{A}_s}\right)_{:,\mathcal{I}_i}\right)^{\dagger}\left(G_{\mathcal{A}_s}^T\right)_{\mathcal{I}_i,:}
    \end{bmatrix}\, .
  \end{equation}
  Next, we insert \Cref{eq:combined_res} into \Cref{sys:4_reordered}
  \begin{equation}
    \label{eq:final}
    \frac{\partial a_s}{\partial a}  = \left(G_{\mathcal{A}_s}\right)_{:,\mathcal{I}_i}\left(\left(G_{\mathcal{A}_s}^T\right)_{\mathcal{I}_i,:}\left(G_{\mathcal{A}_s}\right)_{:,\mathcal{I}_i}\right)^{\dagger}\left(G_{\mathcal{A}_s}^T\right)_{\mathcal{I}_i,:} \, .
  \end{equation}
  The Jacobian is the orthogonal projection onto the column space of $\left(G_{\mathcal{A}_s}\right)_{:,\mathcal{I}_i}$. Despite the non-uniqueness of the solution in \Cref{eq:inactive_solution}, the resulting Jacobian in \Cref{eq:final} is unique \cite[Theorem 1]{Penrose1955}, as the orthogonal projector on a subspace is unique. The rank of an orthogonal projection or projection matrix $P = X\left(X^TX\right)^{\dagger}X^T$ is equal to the design matrix $X$ itself $\text{rank}(\frac{\partial a_s}{\partial a}) = \text{rank}(\left(G_{\mathcal{A}_s}\right)_{:,\mathcal{I}_i})$ \cite[Chapter 9]{Zhang2017}. Due to the shape of $G_{\mathcal{A}_s} \in \mathbb{R}^{d\times n}$, the rank of the Jacobian is at most $d$. However, the columns of $\left(G_{\mathcal{A}_s}\right)_{:,\mathcal{I}_i}$ cannot span all of $\mathbb{R}^d$, since at the optimum the residual $a- a_s$ must be orthogonal to this span. Therefore, the rank is at most $d-1$.
\end{IEEEproof}

\subsection{PROOF OF THEOREM 3: FULL RANK JACOBIAN OF RAY MASK}\label{app:ray_mask_jacobian_rank}
\begin{IEEEproof}
  An easy application and differentiation of the ray mask for $\norm{a - c_{\mathcal{A}_s}} > \epsilon$ is in spherical coordinates, centred at the safe centre. We transform the coordinates with
  \begin{equation}
    a_o = \begin{bmatrix}
      a_{o,r} \\
      a_{o,1} \\
      \vdots  \\
      a_{o,n-1}
    \end{bmatrix} = \text{spherical}(a - c_{\mathcal{A}_s}) \, ,
  \end{equation}
  where we adopt the convention of the radius being the first coordinate. In this coordinate system $c_{\mathcal{A}_s} = 0$, $\lambda_a = a_{o,r}$, and $d_a$ is the first canonical unit vector, which means \Cref{eq:generalised_rm} modifies only the first coordinate
  \begin{equation}
    a_{s, o} =
    \begin{bmatrix}
      \omega(a_{o,r}, \lambda_{\mathcal{A}_s}, \lambda_\mathcal{A})\lambda_{\mathcal{A}_s} \\
      a_{o,1}                                                                              \\
      \vdots                                                                               \\
      a_{o, n-1}
    \end{bmatrix}\, .
  \end{equation}
  Finally, we transform the safe action back
  \begin{equation}
    a_s = \text{spherical}(a_{s, o})^{-1}+c_{\mathcal{A}_s} \, .
  \end{equation}
  The chain rule provides the Jacobian of the ray mask
  \begin{equation}
    \frac{\partial a_s}{\partial a}= \frac{\partial a_s}{\partial a_{s, o}}\frac{\partial a_{s, o}}{\partial a_o}\frac{\partial a_o}{\partial a} \, ,
  \end{equation}
  where the inverse function theorem \cite[Theorem 3.3.2]{ImplicitFunctionTheorem} relates the Jacobians of the transformations as
  \begin{equation}
    \frac{\partial a_s}{\partial a_{s, o}} = \left(\frac{\partial a_o}{\partial a}\right)^{-1} \, .
  \end{equation}
  This relation characterises a similarity transformation, which means $\frac{\partial a_s}{\partial a}$ and $\frac{\partial a_{s, o}}{\partial a_o}$ share the same eigenvalues. The Jacobian $\frac{\partial a_{s, o}}{\partial a_o}$ is
  \begin{equation} \label{eq:ray_gradient}
    \begin{bmatrix}
      \frac{\partial \omega(a_{o,r}, \lambda_{\mathcal{A}_s}, \lambda_\mathcal{A})}{\partial a_{o, r }}\lambda_{\mathcal{A}_s} & \frac{\partial (\omega\lambda_{\mathcal{A}_s})}{\partial a_{o,1}} & \dots & \frac{\partial (\omega \lambda_{\mathcal{A}_s})}{\partial a_{o,n-1}} \\
      \textbf{0}                                                                                                               & \multicolumn{3}{c}{I}                                                                                                                            \\
    \end{bmatrix} \, ,
  \end{equation}
  which is triangular and, therefore, has the eigenvalues in the diagonal elements
  \begin{equation}
    \sigma_i = \begin{cases}
      \frac{\partial \omega(a_{o,r}, \lambda_{\mathcal{A}_s}, \lambda_\mathcal{A})}{\partial a_{o, r }} \lambda_{\mathcal{A}_s} & \text{if } i=1  \\
      1                                                                                                                         & \text{else}\, .
    \end{cases}
  \end{equation}
  Since they are all non-zero, recall the condition on a valid mapping $\frac{\partial w(\lambda_a, \lambda_{\mathcal{A}_s}, \lambda_\mathcal{A})}{\partial \lambda_a} > 0$, and the Jacobian is a square matrix, it has full rank.
\end{IEEEproof}

\subsection{LEARNING CURVES OF ALL LEARNING ALGORITHMS IN NON-SAFEGUARDED TRAINING} \label{app:unsafe_results}
\begin{figure}[H]
  \centering
  \includegraphics[width=\linewidth]{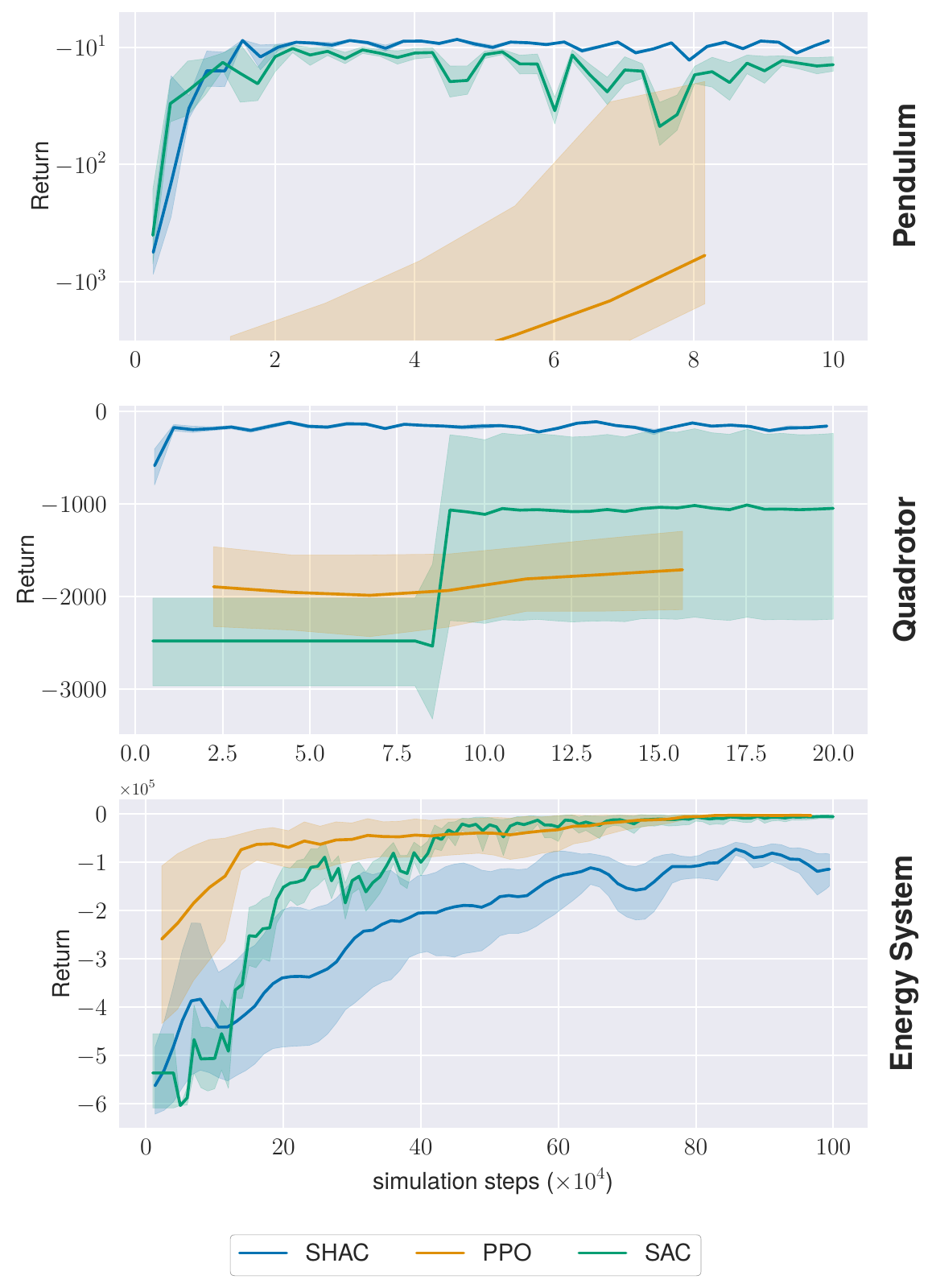}
  \caption{Learning curves of SHAC, SAC, and PPO in non-safeguarded training.}
  \label{fig:unsafe_results}
\end{figure}

\Cref{fig:unsafe_results} shows the performance of SHAC, SAC, and PPO in the absence of any safeguarding. These results complement the main text by illustrating how each algorithm behaves under unconstrained conditions.

\subsection{LEARNING CURVES OF ALL LEARNING ALGORITHMS WITH BOUNDARY PROJECTION} \label{app:results_bp}
\begin{figure}[H]
  \centering
  \includegraphics[width=\linewidth]{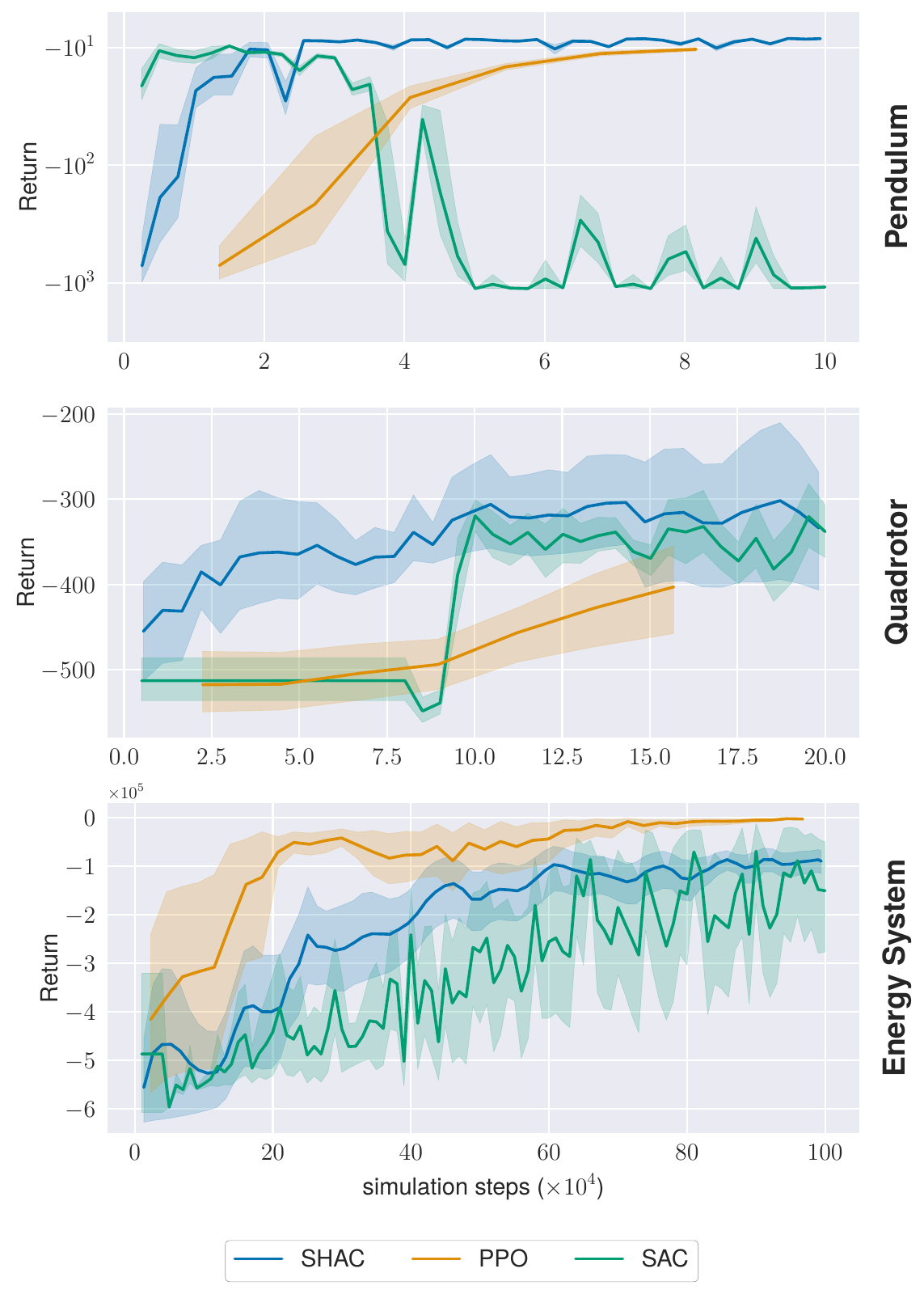}
  \caption{Learning curves of SHAC, SAC, and PPO with boundary projection.}
  \label{fig:results_bp}
\end{figure}

\Cref{fig:results_bp} provides the learning curves for SHAC, SAC, and PPO when applying the boundary projection method. This visualization highlights the effect of the projection on training stability and convergence.

\subsection{LEARNING CURVES OF ALL LEARNING ALGORITHMS WITH RAY MASK} \label{app:results_rm}
\begin{figure}[H]
  \centering
  \includegraphics[width=\linewidth]{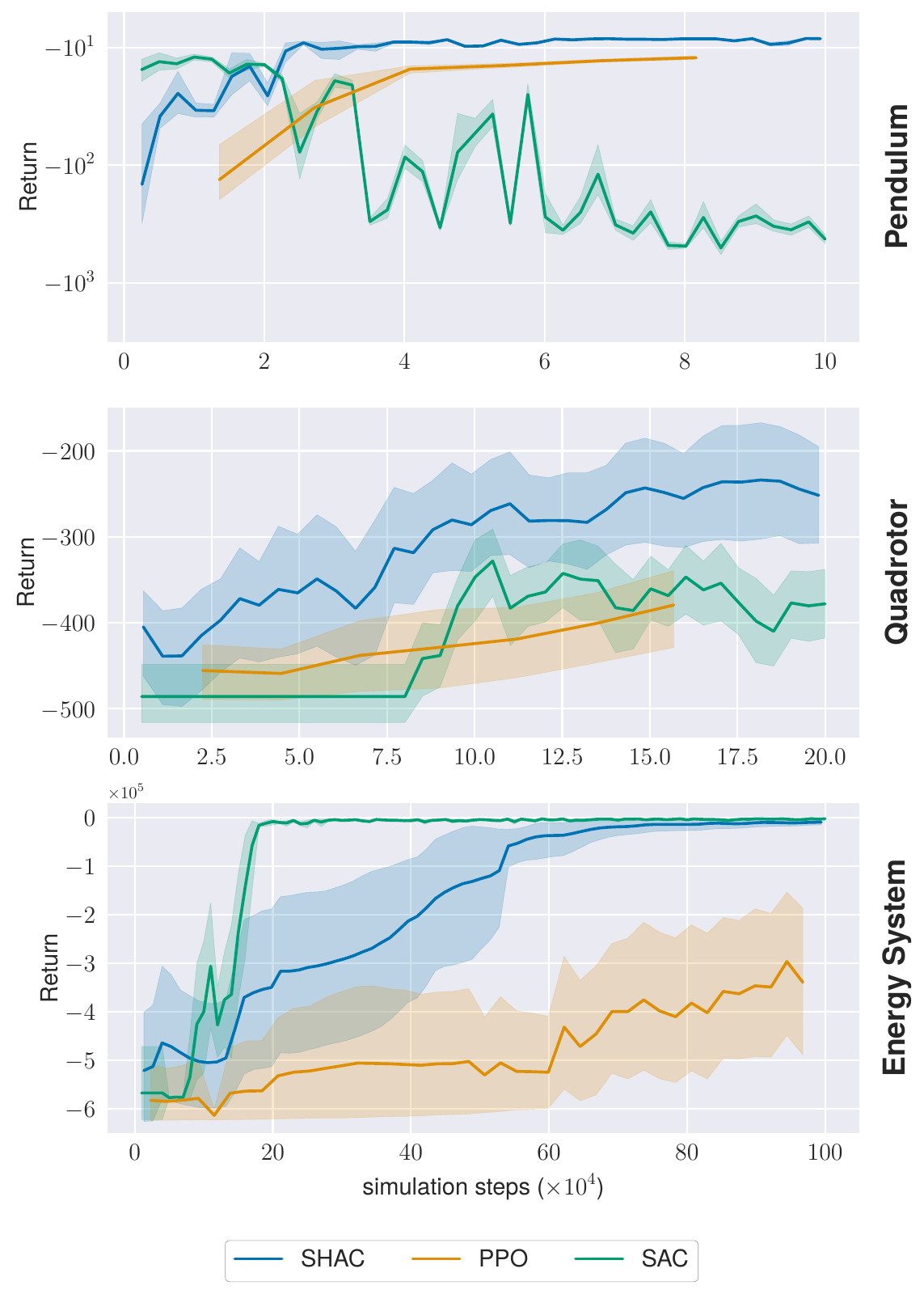}
  \caption{Learning curves SHAC, SAC, and PPO with ray mask.}
  \label{fig:results_rm}
\end{figure}

\Cref{fig:results_rm} presents the learning curves obtained with the ray mask. It illustrates the comparative performance of SHAC, SAC, and PPO when safeguarded by this technique.

\subsection{LEARNING CURVES OF BOTH APPROXIMATIONS} \label{app:orp_vs_zrp}
\begin{figure}[H]
  \centering
  \includegraphics[width=\linewidth]{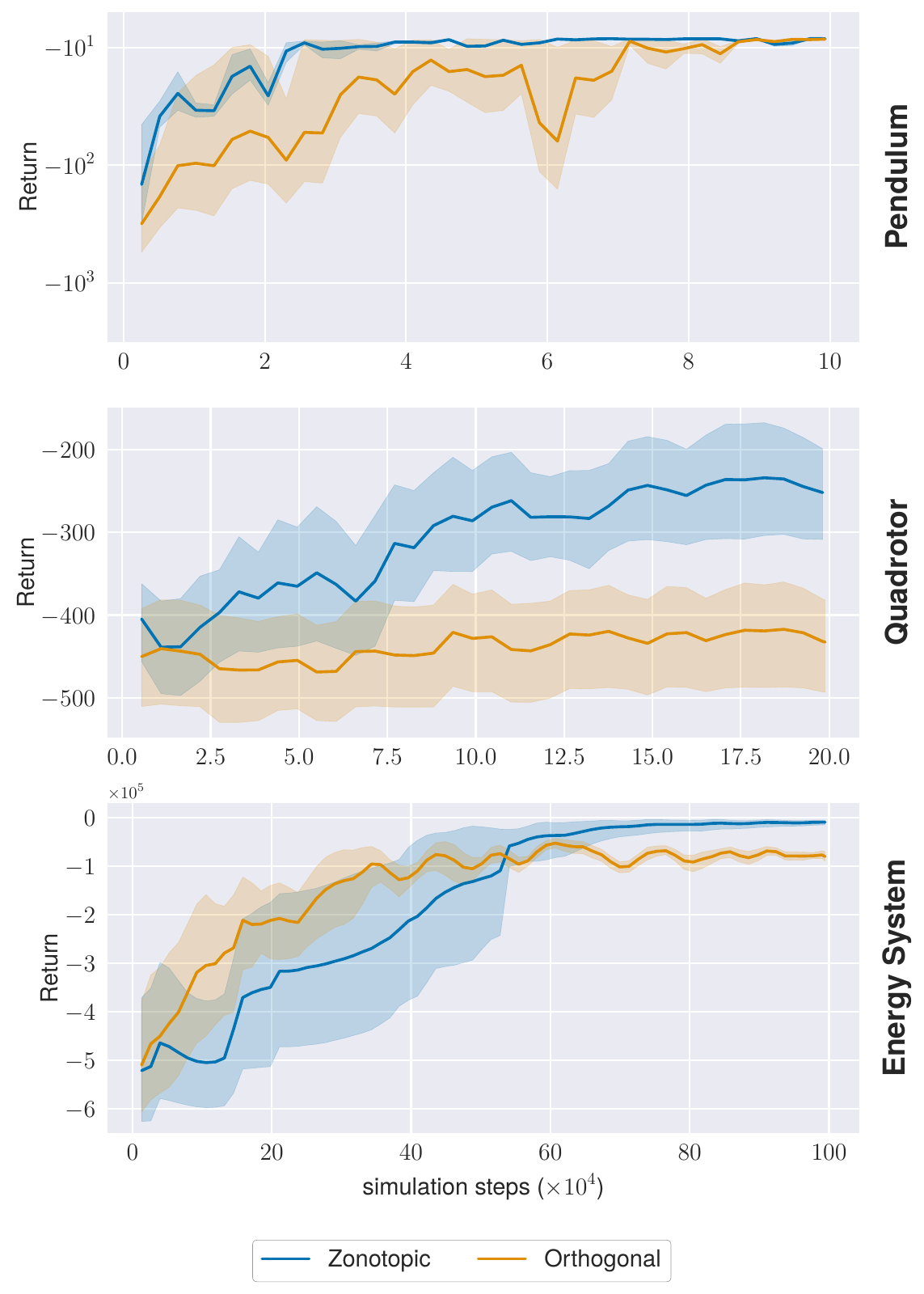}
  \caption{Learning curves of SHAC in safeguarded training with the ray mask, where we compare safe centre approximation methods.}
  \label{fig:orp_vs_zrp}
\end{figure}

\Cref{fig:orp_vs_zrp} compares the safeguarded training results of SHAC under two different safe centre approximation methods. This allows us to assess the relative effectiveness of the ORP and ZRP approximations.

\end{document}